\documentclass[final,1p,times,authoryear]{elsarticle}

\usepackage{amssymb}
\usepackage{amsmath}
\usepackage{amsthm}
\usepackage{csquotes}
\usepackage{xcolor}
\usepackage{float}
\usepackage{booktabs}
\usepackage{subcaption}
\usepackage{longtable}
\usepackage{showlabels}
\usepackage{xurl}
\usepackage[unicode=true,psdextra]{hyperref}

\journal{Journal of Mathematical Psychology}

\begin{document}

\begin{frontmatter}



\title{Ensemble Kalman filter for uncertainty in human language comprehension\tnoteref{t1}} 
\tnotetext[t1]{The research has been funded by the Deutsche Forschungsgemeinschaft (DFG)-   Project-ID 318763901 - SFB1294.}
\author[1]{Diksha Bhandari\corref{corr1}}
\ead{diksha.bhandari@uni-potsdam.de}
\affiliation[1]{organization={Institute of Mathematics, University of Potsdam},
            addressline={Karl-Liebknecht-Str. 24-25},
            city={Potsdam},
            postcode={14476},
            country={Germany}}

             
\author[2]{Alessandro Lopopolo} 
\ead{alessandro.lopopolo@uni-potsdam.de}
\affiliation[2]{organization={Institute of Psychology, University of Potsdam},
            addressline={Karl-Liebknecht-Str. 24-25},
            city={Potsdam},
            postcode={14476},
            country={Germany}}

             
\author[2]{Milena Rabovsky} 
\ead{milena.rabovsky@uni-potsdam.de}
                          
\author[1]{Sebastian Reich}
\ead{sereich@uni-potsdam.de}

\cortext[corr1]{Corresponding author}

\begin{abstract}
Artificial neural networks (ANNs) are widely used in modeling sentence processing but often exhibit deterministic behavior, contrasting with human sentence comprehension, which manages uncertainty during ambiguous or unexpected inputs. This is exemplified by reversal anomalies—sentences with unexpected role reversals that challenge syntax and semantics—highlighting the limitations of traditional ANN models, such as the Sentence Gestalt (SG) Model. To address these limitations, we propose a Bayesian framework for sentence comprehension, applying an extention of the ensemble Kalman filter (EnKF) for Bayesian inference to quantify uncertainty. By framing language comprehension as a Bayesian inverse problem, this approach enhances the SG model’s ability to reflect human sentence processing with respect to the representation of uncertainty. Numerical experiments and comparisons with maximum likelihood estimation (MLE) demonstrate that Bayesian methods improve uncertainty representation, enabling the model to better approximate human cognitive processing when dealing with linguistic ambiguities.
\end{abstract}

%


\end{frontmatter}


\section{Introduction}
\label{sec:intro}
Artificial neural networks (ANNs) have become indispensable tools in modeling sentence processing within the field of natural language processing and cognitive science. These models are capable of handling complex linguistic structures, making accurate predictions, and resolving ambiguities with a notable degree of certainty, even when they are wrong \cite{pmlr-v70-guo17a, Hein_2019_CVPR}. However, this behavior stands in contrast to human sentence comprehension, which often involves managing uncertainty, especially when faced with ambiguous or unexpected language inputs. Human cognition frequently engages in reanalysis, revisions, and flexible interpretations during sentence processing, highlighting a potential mismatch between the deterministic behavior of ANNs and the more adaptive nature of human understanding, which is - among other things - based on representing uncertainty in case of conflicting cues.

This paper addresses the disparity between ANNs and human cognition in processing uncertainty during sentence comprehension, with a particular focus on reversal anomalies -- sentences that disrupt expected meaning patterns. For instance, in a sentence such as ‘‘The dog was bitten by the man" (example 1;\cite{ferreira2003}) or ‘‘Every morning at breakfast the eggs would eat...." (example 2; \cite{kuperberg2003}), typical agent and patient roles are reversed. Reversal anomalies present a significant challenge to both human and artificial systems by introducing conflicts between syntax and semantics that can lead to comprehension breakdowns or reinterpretations. While ANNs such as the Sentence Gestalt Model \cite{sgm1989,STJOHN1990217,rabovsky2018}, can display a high level of certainty even during misinterpretations, human behavioral performance in sentence processing often displays higher levels of uncertainty \cite{ferreira2003}.

To explore this disparity and work towards reducing it, we examine the performance of the Sentence Gestalt Model, an ANN-based model of sentence comprehension, when confronted with reversal anomalies, and compare its handling of uncertainty to a Bayesian approach that is meant to enhance the model's ability to account for uncertainty, which is often overlooked in traditional deep learning models of language processing. In this work, we investigate a sampling method based on an extension of the ensemble Kalman filter (EnKF) for uncertainty quantification. By formulating the task of language comprehension as a Bayesian inverse problem, we aim to improve the model's capacity to reflect the probabilistic nature of human sentence comprehension and resulting uncertainties.

This approach is validated through numerical experiments and a comparison with traditional maximum likelihood estimation (MLE)-based methods. Our results demonstrate that incorporating Bayesian techniques can lead to an enhanced representation of uncertainty in the model when processing ambiguous or conflicting sentence structures, which comes somewhat closer to human cognitive processing of uncertainty. 

\subsection{Uncertainty in human language comprehension}
\label{sec:uq_humans}
Humans, during sentence comprehension, often adopt heuristic strategies to cope with the complexity of language. These strategies enable individuals to generate a \textit{good enough} interpretation of a sentence without engaging in a detailed syntactic analysis. \cite{ferreira2003} suggested that such heuristic approaches prioritize resource efficiency and speed over thoroughness, allowing for rapid comprehension at the expense of occasional misinterpretations, particularly in complex or non-canonical sentences.

These heuristic strategies are reflected in increased behavioral uncertainty, which is measurable through higher error rates and longer reaction times when processing non-canonical sentences, such as reversal anomalies (RA). When confronted with such sentences that involve conflicts between syntactic structure (which would suggest that the man bites the dog in  example 1, above) and plausibility (which would suggest that it should be rather the dog who bites the man), participants are more prone to errors, suggesting that their reliance on heuristic processing is insufficient for correctly interpreting the sentence structure. Specifically, \cite{ferreira2003} reported that about 25 percent of college students misinterpreted simple passive sentences such as example 1 above as indicating that the dog was biting the man, which is incompatible with the sentence's syntax.

RA sentences can be interpreted by relying on either syntactic cues or semantic cues. While syntactic analysis would correctly assign the roles based on sentence structure, semantic cues, such as world knowledge or expectations about event probabilities, often guide interpretation. In RA sentences, humans frequently default to semantic cues, leading to misinterpretations where the brain gravitates toward a more plausible event scenario rather than adhering strictly to syntactic rules as explained for the example above. This tendency can lead to so called good-enough processing and misinterpretations \cite{ferreira2003} and can create what is known in event related potential (ERP) research as a semantic illusion, where the sentence initially seems to make sense based on real-world knowledge (leading to small amplitudes of the N400 ERP component, which is usually large for semantic anomalies, \cite{Kutas2011}), but later requires reanalysis due to its syntactic inconsistency (reflected in large amplitudes of the subsequent P600 component), \cite{kuperberg2003}.

While humans display a high level of uncertainty when interpreting RA sentences due to the conflicting cues, computational models of language processing, such as the SG model originally proposed by \cite{STJOHN1990217} and extended by \cite{rabovsky2018} to account for N400 amplitudes, can exhibit high confidence in their interpretation by consistently favoring semantic cue-based processing of reversal anomalies. In the SG model, the word ``eggs'' in example 2, above, was consistently and confidently interpreted as the patient rather than the agent, reflecting a plausibility-based interpretation that aligns with expected event structures. This confident misinterpretation was likely due to the fact that the SG model had never experienced the eggs as the agents of an eating action during training. Nonetheless, the mismatch with the syntactic structure of the sentence could have been expected to introduce some hints of doubt within the system. The SG model's strong and confident reliance on event probabilities despite conflicting cues underscores how computational systems can be highly confident even when they are wrong, while in  human processing conflicting cues often lead not only to higher error rates but at the same time to greater uncertainty.


\subsection{Uncertainty quantification in artificial neural networks} \label{sec:uq_ann} 
Most ANNs and models in deep learning can be viewed as deterministic functions, providing point estimates of model parameters and predictions. While deep learning models are renowned for their strong generalization capabilities, the growing reliance on these models in modern artificial intelligence applications necessitates not only good generalization but also the ability to recognize when they do not know something \cite{Gal2016UncertaintyID}. As shown in \cite{pmlr-v70-guo17a, Hein_2019_CVPR}, neural networks when fitted by approximating the maximum likelihood estimator (MLE), or, equivalently, minimizing the negative log-likelihood, are not robust to out-of-distribution (OOD) datasets, and do not provide any measure of uncertainty in the model's predictions. In contrast, a Bayesian approach to ANNs enables uncertainty estimates, allowing confidence levels to be assigned to model predictions.

To address the need for uncertainty quantification in ANNs, use of Bayesian neural networks has been introduced in \cite{Gal2016UncertaintyID, neal2011mcmc, neal2012bayesian, dropout_gal16}. In the case of supervised learning, given a set of training inputs and corresponding output labels $ \mathcal{D} = \{x_i, t_i\}^N_{i=1}$, Bayesian inference aims to find the parameters $\theta$ that are \textit{likely} to have generated the given outputs. Following the Bayesian perspective, we put a prior distribution $\pi(\theta)$ over the parameter space, which represents our belief as to which parameters are likely to have generated our output before our model observes any training data. As the model observes more and more data, it refines the prior distribution to learn the posterior distribution. In an attempt to make the techniques of Bayesian inference in deep learning feasible, methods of \enquote{partially} Bayesian neural networks have also been studied in \cite{Daxbergeretal21, Kristiadi2020BeingBE}.

The EnKF \cite{evensen_enkf}, has been a popular choice for performing data assimilation due to its robustness, and ability to handle large-scale and complex dynamical systems. More recently, the EnKF and ensemble Kalman inversion (EKI), have been used as a derivative-free technique for Bayesian inference \cite{Haber2018NeverLB, Kovachki_2019} in neural networks. The EnKF has also been studied as a sampling method for Bayesian inverse problems, focusing on the $l_2$-loss function in \cite{ding2021ensemblesampler}. Such methods transform the prior distribution at $\tau =0$ to samples from the posterior distribution as $\tau \rightarrow \infty$. The work \cite{bhandari2024, pidstrigach2022affine} investigates ensemble transform methods for Bayesian inference in logistic regression, i.e., the case of cross-entropy loss function, within the interactive particle setting.

\subsection{Sentence Gestalt Model}
\label{sec:sgm}
\begin{figure}
    \centering
    \includegraphics[width = 0.9\textwidth]{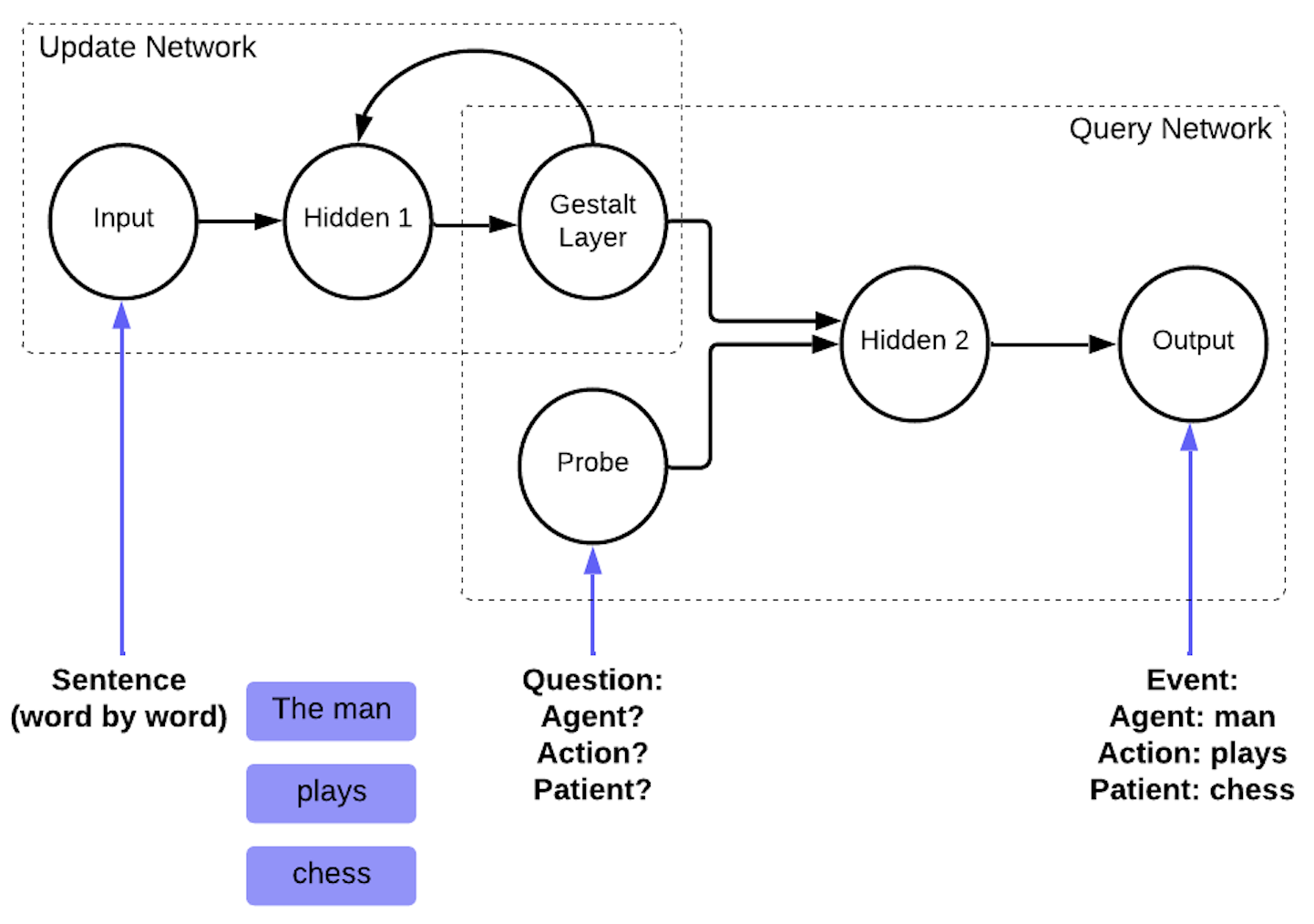}
    \caption{The Sentence Gestalt model}
    \label{fig:sgm_architecture}
\end{figure}

The Sentence Gestalt Model \cite{sgm1989, STJOHN1990217} aims to model the probabilistic aspects of language comprehension by employing a fully connected neural network architecture \cite{rabovsky2018}. The design of the SG model is based on the principle that listeners dynamically update their probabilistic representation of the described event with each incoming word of the sentence. The model is divided into two components: the update network and the query network. The update network functions as a sequential encoder, processing each linguistic constituent in sequence to generate a ``sentence gestalt", which reflects the evolving interpretation of the sentence's meaning. A recurrent layer in the architecture feeds the output from one iteration as input to the next. Each constituent is encoded as an activation pattern over the constituent units, and the gestalt layer captures the model's best current guess of the sentence's meaning. The query network then utilizes this sentence gestalt representation. To decode a specific role-filler pair, the gestalt layer is probed with either the role or filler. The resulting activation from the probe combines with the sentence gestalt in a second hidden layer, producing the activation of the corresponding role-filler pair in the output layer. The architecture of the SGM is illustrated in Figure \ref{fig:sgm_architecture}. In this study, we compare a standard SG model with its Bayesian implementation to examine whether the latter improves the model's ability to represent uncertainty.

The structure of this paper is as follows. Section \ref{sec:methods} provides the necessary mathematical background on ensemble transform methods for Bayesian inference in logistic regression. Additionally, it details the formulation and implementation of the proposed sampling method for a Bayesian Sentence Gestalt model, incorporating dropout techniques and an efficient time-stepping scheme. We present and evaluate numerical results comparing the proposed Bayesian SG model with its standard maximum likelihood estimation (MLE)-based implementation in Section \ref{sec:results}. Section \ref{sec:discussion} offers a discussion of the findings, and concludes with a summary and final remarks.

\section{Materials and methods}
\label{sec:methods}
As discussed in Section \ref{sec:uq_ann}, an MLE training scheme is not robust to out-of-distribution (OOD) dataset, and does not provide any measure of uncertainty in the model's predictions. To overcome this problem, we propose a Bayesian inference approach by replacing the point estimator $\hat{\theta}$ with a posterior distribution $\pi(\theta)$ over the model parameters. Adopting a Bayesian approach, we place a prior distribution $\pi_{\text{prior}}(\theta)$ on the output layer parameters and refine it with data to learn the posterior distribution $\pi_{\text{post}}$. Using Bayes' theorem, the posterior distribution is given by
\begin{equation}\label{eqn:bayes}
\pi_{\text{post}}(\theta) \propto \exp(-l(\theta)) \,\pi_{\text{prior}}(\theta),
\end{equation}
where $l(\theta)$ denotes the negative log-likelihood function.

Consider the data set $\mathcal{D} = \{(\psi^n, t^n)\}_{n = 1}^{N},$ where $t^n$ are targets for the binary classification with input features $\psi^n \in \mathbb{R}^D \text{ for } n=1,...,N$. The $\psi^n$'s can either be the representation of the data $(x^n, t^n)$ in any numerical form, or stem from a feature map. In this work, we will train the SGM with MLE approach for use as a feature map in Bayesian inference. We decompose the $r$-layered SGM into a feature map $\psi_{\hat{\theta}}: \mathbb{R}^n \rightarrow \mathbb{R}^d$, representing the first $r-1$ layers, and an output layer. The output layer is expressed as:
\begin{equation}
F_{\tilde{\theta}}(x) = \sigma(\langle \theta, \psi_{\hat{\theta}}(x) \rangle),
\end{equation}
where $\tilde{\theta} = (\hat{\theta}^{\rm T}, \theta^{\rm T})^{\rm T}$. We first train the entire network using an MLE approach to obtain $\tilde{\theta}_{\text{MLE}}$ and the trained feature map $\psi(x) = \psi_{\hat{\theta}_{\text{MLE}}}(x)$.
\begin{figure}
    \centering
    \includegraphics[width= 0.9\textwidth]{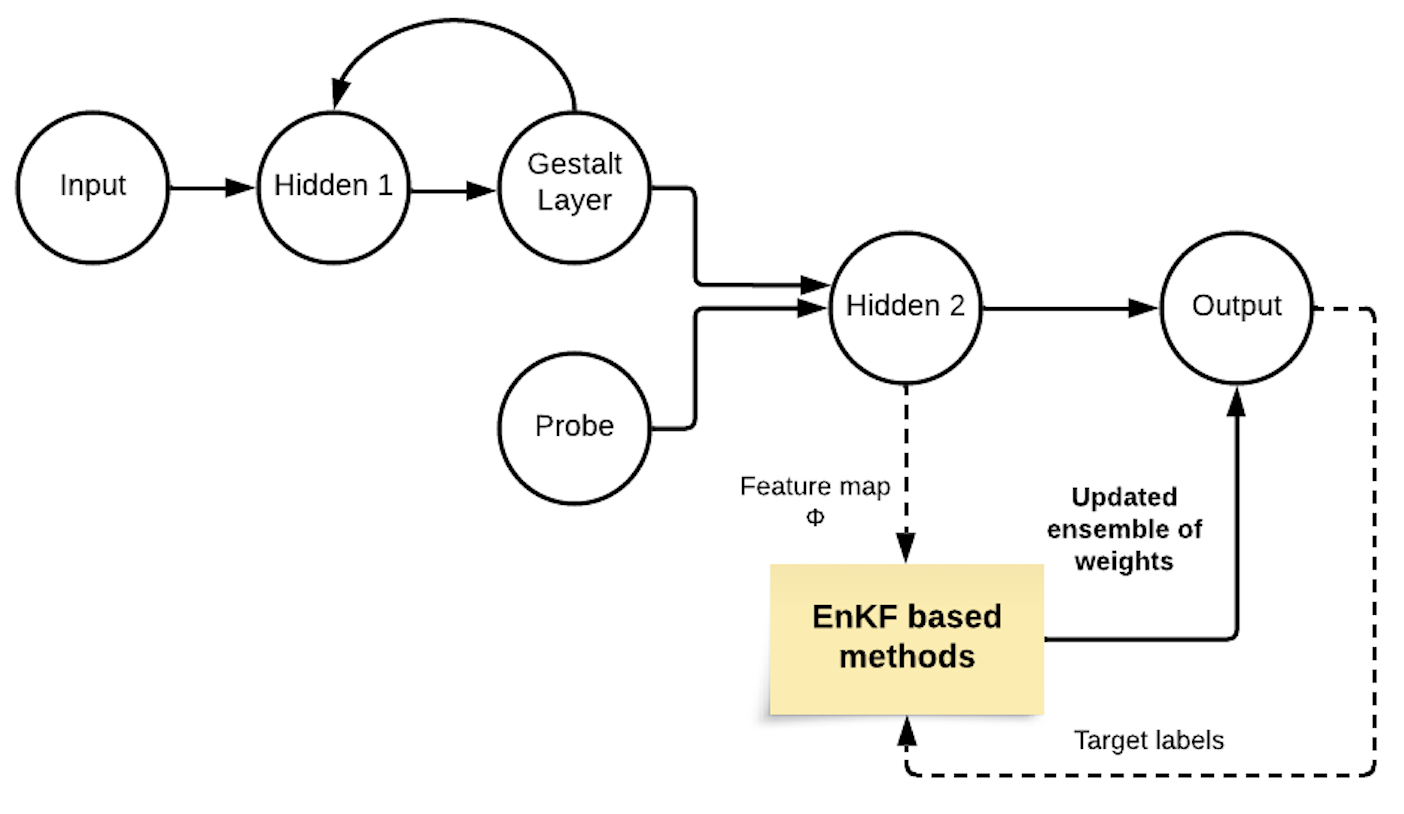}
    \caption{Last-layer Bayesian inference in SGM using ensemble transform methods}
    \label{fig:ll_enkf_sgm}
\end{figure}
So, after observing data $\mathcal{D}$, the goal is to infer $\pi (\theta | \mathcal{D} )$ using Bayes' formula \eqref{eqn:bayes}. The predictive distribution in the case of binary classification is given by
\begin{equation}\label{eqn:pred_dist}
    \pi (t=1 |x, \mathcal{D} ) = \int \sigma (F_{\theta}(x) \pi(\theta | \mathcal{D}) \rm d\theta,
\end{equation}
where $\sigma$ is the sigmoid function, defined as 
\begin{equation}\label{eqn:sigmoid}
\sigma(a) = \frac{1}{1 + \exp(-a)}.
\end{equation}
Since, the above integral \eqref{eqn:pred_dist} is intractable, we use the ensemble of particles distributed according to $\pi_{\text{post}}$, approximated with an EnKF-based method of Bayesian inference to estimate the predictive distribution as 
\begin{equation}\label{eqn:pred_ensemble}
\pi(t=1|x, \mathcal{D}) = \frac{1}{J} \sum_{j=1}^J \sigma(\langle \theta_j^*, \psi_{\hat{\theta}_{\text{MLE}}}(x) \rangle),
\end{equation}
where $\{\theta_j^*\}_{j=1}^J$ represents the final ensemble of particles. Figure \ref{fig:ll_enkf_sgm} illustrates the implementation of a last-layer Bayesian inference in the SGM using EnKF-based methods. In the forthcoming sections, we describe an ensemble transform method to infer $\theta_j^*$, such that the predictive distribution \eqref{eqn:pred_dist} can be approximated by \eqref{eqn:pred_ensemble} in a tractable and efficient manner. 

\subsection{Ensemble transform methods for Bayesian logistic regression}
\label{sec:enkf_lr}
For binary classification, we consider the dataset $\mathcal{D} = \{(\psi^n, t^n)\}_{n=1}^N$, where $\psi^n \in \mathbb{R}^D$ are input features, and $t^n \in \{0, 1\}$ are binary targets. The probability of class membership is given by
\begin{equation}\label{eqn:y}
P_{\theta}[\psi \in C_1] := y_n(\theta) = \sigma(\langle \theta, \psi \rangle),
\end{equation}
where $\sigma$ is the sigmoid function defined in \eqref{eqn:sigmoid}, and $\langle \cdot, \cdot \rangle$ is the inner product between two vectors. Further, we introduce the notation
$$\Psi = (\psi^1,...,\psi^N) \in \mathbb R ^{\mathit {D \times N}}.$$
The negative log-likelihood is given by cross entropy loss function
\begin{equation}\label{eqn:cee}
l(\theta) = -\sum_{n=1}^N \left( t_n \log(y_n(\theta)) + (1 - t_n) \log(1 - y_n(\theta)) \right).
\end{equation}
The gradient of the negative log-likelihood $l$ is given by
\begin{equation}
    \nabla_\theta l(\theta)= \sum_{n=1}^N (y_n(\theta) - t^n)\psi^n = \Psi(y(\theta) - t),
    \label{eqn:grad_Psi}
\end{equation}
and the Hessian is
\begin{equation}
    D_\theta^2 l(\theta) = \Psi R(\theta) \Psi^{\rm T}.
    \label{eqn:hessian_Psi}
\end{equation}
Here $R(\theta) \in \mathbb{R}^{N \times N}$ is a diagonal matrix with diagonal entries 
\begin{equation*}
    r_{nn} = y_n(\theta)(1 - y_n(\theta)).
    \label{eqn:def_R}
\end{equation*}
We denote by $\theta^1_{\tau}, \ldots \theta^J_{\tau}$ an ensemble of $J$ particles, indexed by a time parameter $\tau \geq 0$. We denote by $m_{\theta_{\tau}}$ the empirical mean of the ensemble,
\begin{equation}\label{eqn:emp_mean}
	m_{\theta_{\tau}} = \frac{1}{J} \sum_{j=1}^J \theta_{\tau}^{j},
\end{equation}
and by $P_{\theta_{\tau}}$ the empirical covariance matrix,
\begin{equation}\label{eqn:emp_cov}
	P_ {\theta_{\tau}} = \frac{1}{J-1} \sum_{j=1}^{J}(\theta_{\tau}^{j}- m_{\tau} ) (\theta_ {\tau}^{j}- m_{\tau})^{\rm T}.
\end{equation} 
We introduce the matrix of ensemble deviations $\Theta_s \in \mathbb R^{D \times J}$ as
\begin{equation}\label{eqn:ens_dev}
    \Theta_{\tau} = \left(\theta_{\tau}^{1}-m_{\theta_{\tau}},\theta_{\tau}^{2}-m_{\theta_{\tau}},\ldots,\theta_{\tau}^{J}-m_{\theta_{\tau}}  \right).
\end{equation}
We will adapt the convention that upper indices stand for the ensemble index, while lower indices stand for the time index or the $i$th entry of an $N$-dimensional vector.
Associated to the particle system is the empirical measure
\begin{equation*}
	\mu_{\theta_{\tau}} = \frac{1}{J} \sum_{j=1}^J \delta_{\theta^j_{\tau}},
\end{equation*}
where $\delta_\theta$ stands for the Dirac delta at $\theta$. We denote the expectation with respect to this measure by $\mu_{\theta_{\tau}}[g]$, i.e.~for a function $g$,
\[
	\mu_{\theta_{\tau}}[g] = \frac{1}{J} \sum_{j=1}^J g(\theta^j_{\tau}).
\]
In the following subsection, we briefly describe the dynamical system formulation of an ensemble transform method for Bayesian inference in a logistic regression problem. For further details on the derivation of the method, refer \cite{bhandari2024}.

\subsubsection{Dropout deterministic sampler}
\label{sec:infinite_det_sampler}
The dropout deterministic sampler presented in this section operates by simulating a dynamical system where an ensemble of parameters $\{\theta_{\tau}^j\}_{j=1}^J$ evolves over time $(\tau \geq 0)$ according to an evolution equation described by an interacting particle system (IPS) ODE. Starting from the initial ensemble $\{\theta_{0}^j\}_{j=1}^J \sim \pi_{\rm prior}$, the IPS is structured to harness the evolution of the empirical mean and covariance, guiding the ensemble towards regions of high posterior probability, as the model sees more and more data. The system then samples from the target distribution in the limit $\tau \to \infty$.

The key idea of the deterministic (second-order)  sampler introduced in \cite{bhandari2024} is to combine the homotopy approach \cite{pidstrigach2022affine} with a deterministic equivalent of overdamped Langevin dynamics to motivate an IPS that approximates the posterior. Therefore, one can construct systems that sample the target distribution as $\tau \to \infty$, a property of the most famously used Markov Chain Monte Carlo (MCMC) algorithms. However, the computational complexity of these MCMC methods makes them impractical to use in Bayesian inference for large ANNs. Thus, to make the Bayesian approach feasible, we use the deterministic second-order sampler introduced in \cite{bhandari2024}, which has been shown to demonstrate similar uncertainty estimates as the Hamiltonian Monte Carlo (HMC) method. The advantage of using the deterministic sampler is that it has been experimentally shown to have  faster convergence towards equilibrium than HMC, and the particles stop moving once they reach equilibrium. Moreover, under appropriate conditions, the determinsitic sampler gives an exact representation for the evolution of mean $m_{\theta_{\tau}}$ and covariance $P_{\theta_{\tau}}$ in the limit $\tau \rightarrow \infty$. The IPS ODE we will use is given by
\begin{equation}\label{eqn:sampler_ode}
\begin{split}
\frac{\mathrm{d} \theta_{\tau}^j}{\mathrm{d} \tau}  & =  -\frac{1}{2}P_{\theta_{\tau}} \Psi \Bigl( \mu_{\theta_{\tau}}[R] \Psi^{\rm T}(\theta_{\tau}^j -                                                     m_{\theta_{\tau}}) + 2 (\mu_{\theta_{\tau}}[y] - t) \Bigr) \\
                                         & \qquad  -\,\frac{1}{2}P_{\theta_{\tau}} P_{\rm{prior}}^{-1}(\theta^j_{\tau} + m_{\theta_{\tau}} - 2 m_{\rm{prior}})
                                         + \frac{1}{2}(\theta^j_{\tau} -m_{\theta_{\tau}}) 
\end{split}
\end{equation}
with random initial conditions $ \theta^j_0 \sim \mathcal{N}(m_{\rm{prior}}, P_{\rm{prior}})$ i.i.d. for $j = 1, \ldots, J$. In Section \ref{sec:dropout_stepping}, we describe the dropout technique and linearly implicit, efficient methods of time-stepping, used in combination with the deterministic sampler for robust numerical implementation. 

\subsection{Dropout and time-stepping}
\label{sec:dropout_stepping}
For ensemble sizes $K \leq J$, the ensemble $\{\theta_ {\tau}^{j}\}_{j=1}^K$ tends to propagate in the subspace defined by the initial ensemble parameters. To overcome this, we use ensemble dropout, where a fraction $\rho \in (0, 1)$ of the entries in $\{\theta_ {\tau}^{j}\}, j=1...,K$ are randomly set to zero. The dropout technique, first introduced for neural networks in \cite{dropout_srivastava14a}, can be viewed as an implicit regularization method to avoid overfitting during training. For high dimensional Bayesian logistic regression problems, the dropout in our ensemble transform method can be intercepted as a similar technique to avoid spurious correlations, and break the ``subspace property" in the small ensemble size setting \cite{pidstrigach2022affine}. Thus, the empirical covariance matrix is modified as
\begin{equation}\label{eqn:dropout_cov}
\hat{P}_{\theta_{\tau}} = \frac{1}{(1 - \rho)(J - 1)} \hat{\Theta}_{\tau} \hat{\Theta}_{\tau}^{\rm T},
\end{equation}
where $\hat{\Theta}_s$ represents the ensemble deviations \eqref{eqn:ens_dev} after dropout.

 Now, we introduce a step-size $\Delta \tau \geq 0$ and discrete times ${\tau}_s = s \Delta \tau$. In this section, we will use the shorthand $\theta_{\tau_s} \approx \theta_s, m_{\theta_{\tau_{s}}} \approx m_{s}, P_{\theta_{\tau_{s}}} \approx P_{s}, \Theta_{\tau_{s}} \approx \Theta_{s},  \mu_{\theta_{\tau_{s}}} \approx \mu_{s}$. We use a Trotter splitting approach to numerically solve \eqref{eqn:sampler_ode}, for details see \cite[Section 4.2]{bhandari2024}.
Therefore, for every time step $s$, given $\{\theta_{s}^j\}_{j=1}^J$, we compute $\theta_{s+1/2}^j$ using the homotopy-based moment matching, with a tamed method of discretization given by
\begin{equation}\label{eqn:homotopy_stepping}
    \theta_{s+1/2}^j = \theta_s^j - \frac{\Delta \tau}{2} \hat{P}_s \Psi \left( M_s \Psi^{\rm T} (\theta_s^j -m_s) + 2( \mu_{s}[y] - t ) \right)
\end{equation}
where $\hat{P}_s$ is the dropout empirical covariance matrix defined in \eqref{eqn:dropout_cov}, and
\begin{equation}\label{eqn:K}
M_{s} = \left(\Delta \tau \Psi^{\rm T}\hat{P}_{s} \Psi +  \mu_{s} [R]\right)^{-1}
\end{equation}
for $j = 1, \ldots, J$. For the second half step, we apply the following discretization scheme
\begin{equation}\label{eqn:sampler_stepping}
\begin{split}
    \theta_{s+1}^{j} & = \theta_{s+1/2}^{j} -\frac{\Delta \tau}{2}  \hat{P}_{s+1/2} \left(\Delta \tau \hat{P}_{s+1/2} + P_{\rm{prior}} \right)^{-1} (\theta_{s+1/2}^{j} +  m_{s+1/2} -2 m_{\rm{prior}})  \\
    & \quad +  \frac{\Delta \tau}{2}(\theta^j_{s+1/2} -m_{s+1/2}).
\end{split}
\end{equation}
Note that, since $P_{\rm prior}$ is diagonal, full matrix inversion in (\ref{eqn:sampler_stepping}) is replaced by inverting the diagonal only.

We provide a pseudo-code describing the algorithm for dropout deterministic sampler in Algorithm \ref{table:algorithm_sampler}.
\begin{table}[H]
\centering
\refstepcounter{table}\label{table:algorithm_sampler}
\begin{tabular}{p{\textwidth}}
\hline 
\textbf{Algorithm 1: Dropout deterministic sampler for Bayesian inference in SG Model} \\ [0.5ex] 
\specialrule{1.5pt}{0pt}{0pt}
\vspace{0.1ex}
\textbf{Inputs}: Data set $\{(\psi^n, t^n)\}_{n = 1}^{N}$; 
 feature map $\Psi = \{\psi^n\}_{n = 1}^{N}$;
 initial ensemble $\{\theta_{0}^{j}\}_{j = 1}^{J}$; dropout rate $\rho$; step-size $\Delta \tau$; threshold value $\epsilon > 0$.\\[1ex]

\textbf{Initialize}: $s = 0$. \\

\textbf{while} $ \frac{\| \hat{P}_{s+1} - \hat{P}_s \|_2}{\| \hat{P}_s \|_2} \geq \epsilon$ \textbf{do:} \\
\hfill\begin{minipage}{\dimexpr\textwidth-1mm}
    \begin{enumerate}
        \item Numerically solve ODE \eqref{eqn:sampler_ode} by trotter splitting:
            \begin{enumerate}
                \item Compute ensemble mean $m_s$ \eqref{eqn:emp_mean}, dropout ensemble deviations $\hat{\Theta}_s$, and dropout covariance $\hat{P}_s$ \eqref{eqn:dropout_cov}.
                \item Evaluate  $y$ \eqref{eqn:y}, $\mu_{s}[y]$, $\mu_{s}[R]$ and $M_s$ \eqref{eqn:K}.
                \item Determine $\{\theta^j_{s+1/2}\}_{j=1}^J$ by evolving $\theta_{s+1/2}$ using the time-stepping  \eqref{eqn:homotopy_stepping}.
                \item Evaluate dropout covariance matrix $\hat{P}_{s+1/2}$ \eqref{eqn:dropout_cov}.
                \item Determine $\{\theta^j_{s+1}\}_{j=1}^J$ using the time-stepping (\ref{eqn:sampler_stepping}).
            \end{enumerate}
        \item Update the ensemble $\{\theta_{s}^j\}_{j=1}^J \rightarrow \{\theta_{s+1}^j\}_{j=1}^J$.
    \end{enumerate}
    Increment $s$.
    \vspace{0.6em}
\end{minipage}
\textbf{end while}\\
\textbf{Output:} Final ensemble $\{\theta^{j}_s \}_{j=1}^J$.\\[1ex]
\hline
\end{tabular}
\end{table}

\subsection{Training corpus}
\label{sec:corpus}
Model training and parameter estimation were performed on a synthetically generated corpus consisting of $N=10,000$ sentences made of 3-5 words.  Similar to a previous implementation \cite{rabovsky2018},  events are described in terms of an action, an agent, a location, a situation, and a patient or object onto which the action is performed, and were composed of active transitive structures. The corpus contained 72 unique tokens \textit{(see \ref{sec:corpus_words} for a complete list of words that form the input constituents)}, representing a diverse set of concrete nouns and action verbs. The corpus consists of 12 actions, one out of which was randomly chosen with equal probability. Then, an appropriate agent out of the 4 possible options (man, woman, boy, and girl with different probabilities) was chosen to perform the activated action. This was followed by choosing a  suitable situation for the action. An action can occur without any situation, some actions in one, and some with two possible situations. Then, depending on the action, a patient (out of the 36 objects) was selected.  Depending on the action and situation, a location may or may not be present in the sentence. Thus, possible sentence structures are: (i) for sentences with situation, [situation][agent][action][patient][location], and (ii) for sentences without situation, [agent][action][patient][location, if any]. 
Each word in the corpus was encoded using 176 binary semantic features, hand-coded to capture key semantic properties. As described above, binary semantic features were used as filler-concept representation and therefore presented to the model as probe (to the probe layer) and target (to the output layer) during training. Table \ref{tab:sample_semantic_representations} shows an example of words and their semantic representations used to construct the training environment for the SG model.

\begin{table}
\centering
\begin{tabular}{p{0.15\textwidth} p{0.7\textwidth} } 
\hline 
\textbf{Words} & \textbf{Semantic representations} \\
\hline
woman & agent, adult, female, woman \\
eat  & action, consume, done with foods, eat \\
kitchen & location, inside, place to eat, kitchen \\
breakfast & situation, food related, in the morning, breakfast \\
egg  & consumable, food, white, egg \\
during/at & no output units (activated together with situation words, e.g., ``at breakfast") \\
\hline
\end{tabular}
\caption{Sample words (inputs) and their semantic representations (probes and outputs).}
\label{tab:sample_semantic_representations}
\end{table}

\subsection{Model training with maximum likelihood estimation}
\label{sec:mle} 
The input dataset comprises a randomly sampled training input of (constituent, probe) pairs described above, denoted by $x_{n}= (i_{n}, p_{n}), n = 1, \dots , N$, as a combined sequential input data. The target (output) labels are possible responses to probes being 1 (if the feature is present) or 0 (if the feature is not present). A fully-connected layer followed by a sigmoid activation function is applied to the output of the neural network and the result is compared to the given target labels using a cross entropy loss function. We first train the complete SG model $F_{\theta}(x)$ with a regularized MLE technique, using ADAM optimizer with a learning rate of $4 \times 10^{-3}$ and weight decay for optimization. The model is run for 60 epochs using cross validation and an ``early stopping" criteria with a goal of monitored training to minimize the cross entropy loss and avoid overfitting during training. The MLE approach  minimizes the cross entropy loss \eqref{eqn:cee} across the training set, which provides the minimizer $\theta_{\text{MLE}}$ and an associated feature map $\psi(x):= \psi_{\theta_{\text{MLE}}}(x)$.

\subsection{Experimental conditions}
\label{sec:exp_conditions}
We performed two types of analyses on each comparison: a model analysis and an item analysis, both evaluated using a set of eight semantically congruent (canonical) sentences, each paired with its corresponding reversal anomaly (RA) version based on ten runs of the model. In the congruent sentences, the scenarios depicted plausible agents performing typical actions relevant to the context (e.g., ``During dinner, the woman eats$\dots$"). For the reversal anomalies, each scenario was modified by replacing the agent with a high-probability patient related to the context, followed by an action normally associated with the agent (e.g., ``During dinner, the pizza eats$\dots$"). A model analysis was performed using output values averaged over items within each version for $m=10$ models with different random initializations. An item analysis was performed with values averaged over models and for each item ($n=8$ items).

\section{Results}
\label{sec:results}
\begin{figure}
    \centering
    \begin{subfigure}[b]{0.495\textwidth}  
        \centering
        \caption{ADAM train and test loss}
        \includegraphics[width=\textwidth]{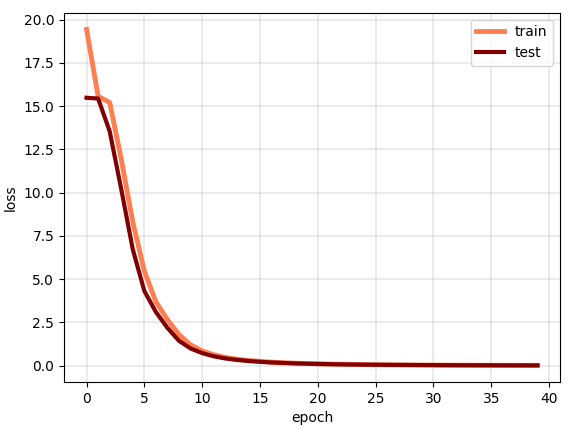} 
        \label{fig:adam_loss}
    \end{subfigure}
    \hfill
    \begin{subfigure}[b]{0.495\textwidth} 
        \centering
        \caption{ADAM train and test accuracy}
        \includegraphics[width=\textwidth]{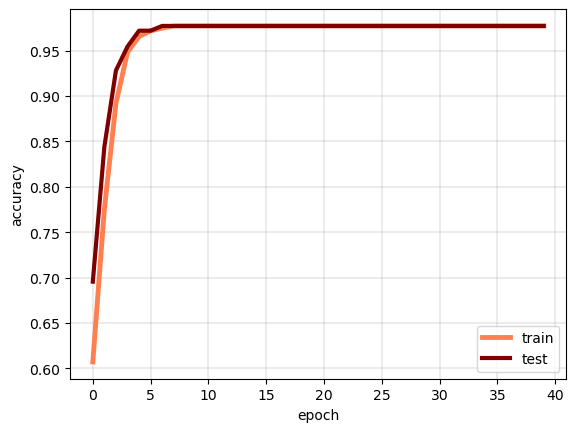} 
        \label{fig:adam_accuracy}
    \end{subfigure}
    \caption{Training \& test loss and accuracy obtained by training SGM with ADAM optimizer.}
    \label{fig:adam_loss&accuracy}
\end{figure}

In this section, we present numerical and statistical results for output activations of the SG model, when trained with standard MLE estimates, and Bayesian approximation with the proposed EnKF-based dropout deterministic sampler \ref{table:algorithm_sampler}. The training set $\{(x^n, t^n)\}_{n = 1}^{N}$ consists of sentences and queries about the events the sentences describe (inputs) so that it can provide responses (labels) to such queries. We train the complete SG model using a regularized MLE technique as described in Section \ref{sec:mle}. The model has a test accuracy of $97.84\%$. Figure \ref{fig:adam_loss&accuracy} shows the plots for model loss and accuracy during train and test phases.

Training using the MLE approach provides the minimizer $\theta_{\text{MLE}}$ and an associated feature map $\psi_{\text{MLE}}(x)$. This feature map $\psi_{\text{MLE}}(x)$ is used for Bayesian logistic regression over multiple output features. We implement the proposed dropout deterministic sampler for Bayesian inference using a Gaussian prior with mean $m_{\text{prior}}= \theta_{\text{MLE}}$ and covariance matrix $P_{\text{prior}} = I$. The results reported in this section are obtained using a dropout rate $\rho = 0.3$ and $\tau = 17$ discrete time-steps. We use the final ensemble of particles $\{\theta^j_{\ast}\}_{j=1}^J$ distributed according to $\pi_{\text{post}}(\theta)$, approximated with the proposed sampler to estimate the predictive distribution as described in \eqref{eqn:pred_ensemble}.

\subsection{Processing congruent and reversal anomaly sentences} 
\label{sec:output_activations}
In this experiment, we report the activation of selected output units when the model is presented with test sentences consisting of semantically congruent and their correspondent RA versions. For both cases, we examine the mean activations of the selected output units across all test sentences ($n=8$ items) and compare the probability of feature activation using the MLE-trained and the Bayesian SG model.  

\begin{figure}
\parbox{\textwidth}{\textbf{\enquote{During dinner, woman eats pizza}}}
    \centering
    \begin{subfigure}[b]{0.45\textwidth}
        \centering
        \includegraphics[width=\textwidth]{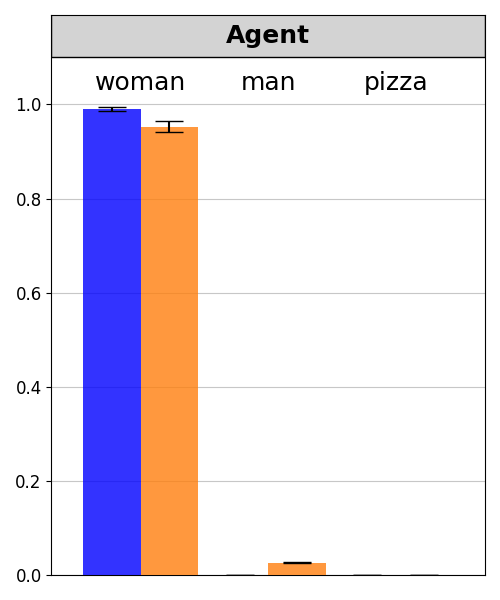}
    \end{subfigure}
    \begin{subfigure}[b]{0.45\textwidth}
        \centering
        \includegraphics[width=\textwidth]{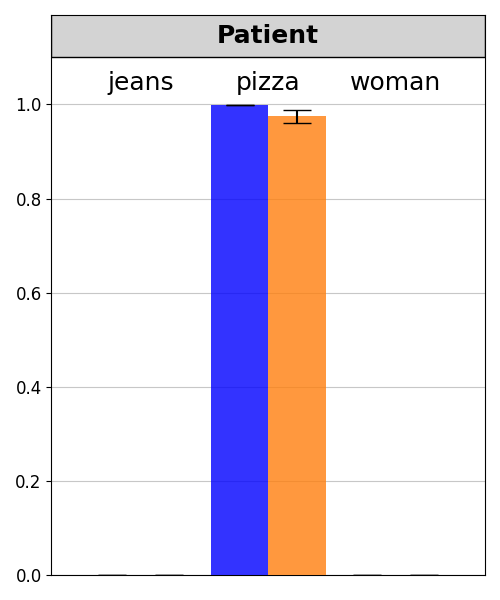}
    \end{subfigure}
    
    \begin{subfigure}[b]{0.55\textwidth}
        \begin{center}
        \includegraphics[width=\textwidth]{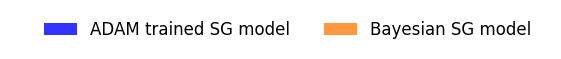}
        \end{center}
    \end{subfigure}
\caption{Mean activations (with standard deviations) of selected output units for the congruent form of test sentences using the ADAM-trained SG model and the Bayesian SG model (with dropout deterministic sampler). Blue bars indicate the output activations of the selected feature (word) from the ADAM-trained model, while red bars represent the predicted probability of the feature being activated in the Bayesian SG model. Black error bars denote standard deviations.}
\label{fig:congruent_mle_sampler}
\end{figure}

\begin{figure}
\parbox{\textwidth}{\textbf{\enquote{During dinner, pizza eats woman}}}
    \centering
    \begin{subfigure}[b]{0.45\textwidth}
        \centering
        \includegraphics[width=\textwidth]{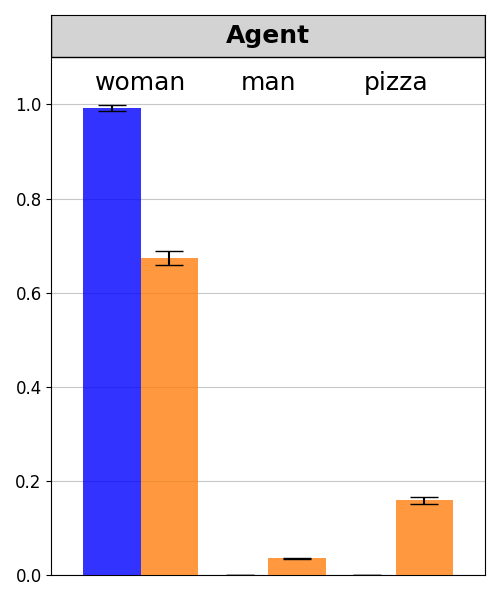}
    \end{subfigure}
    \begin{subfigure}[b]{0.45\textwidth}
        \centering
        \includegraphics[width=\textwidth]{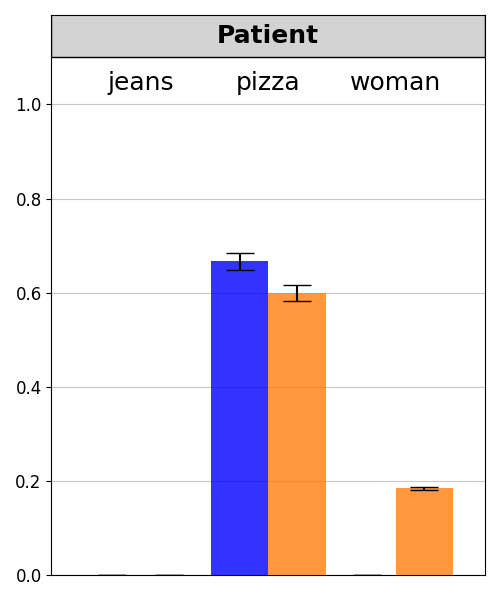}
    \end{subfigure}
    \begin{subfigure}[b]{0.55\textwidth}
        \centering
        \includegraphics[width=\textwidth]{main_legend.png}
    \end{subfigure}
\caption{Mean activations (with standard deviations) of selected output units for the RA form of test sentences using the ADAM-trained SG model and the Bayesian SG model (with dropout deterministic sampler). Blue bars indicate the output activations of the selected feature (word) from the ADAM-trained model, while red bars represent the predicted probability of the feature being activated in the Bayesian SG model. Black error bars denote standard deviations.}
\label{fig:ra_mle_sampler}
\end{figure}

We report and compare the  the mean activations of the output units for congruent form of test sentences using the ADAM-trained SG model and the Bayesian SG model in Figure \ref{fig:congruent_mle_sampler}. Additionally, mean activations for RA test sentences are shown in Figure \ref{fig:ra_mle_sampler}. Mean activations obtained after each word in the test sentences is presented to the model are provided in the Appendix, with details on processing of congruent and RA sentences using the ADAM-trained SG model shown in Figures \ref{fig:unviolated_mle} and \ref{fig:violated_mle}, respectively. 
As shown in Figure \ref{fig:congruent_mle_sampler}, the MLE-trained SG model exhibits high mean output activations for semantically plausible agents and patients in congruent sentences. For instance, the model strongly activates ‘‘woman" when probed for the agent and ‘‘pizza" when probed for the patient in the sentence. When presented with the RA versions of these test sentences, as illustrated in Figure \ref{fig:ra_mle_sampler}, the MLE-trained model maintains high mean output activations for semantically plausible agents and patients. For example, the model continues to output ‘‘woman" as the agent and ‘‘pizza" as the patient in RA sentences, demonstrating a strong reliance on semantic cues even in reversal anomaly contexts.

The performance of the Bayesian SG model, implemented using the dropout deterministic sampler, was compared to the traditional MLE-trained SG model. As shown in Figure \ref{fig:congruent_mle_sampler}, the Bayesian SG model achieves comparable mean activations for semantically plausible agents and patients in congruent sentences. However, when handling RA sentences, the Bayesian model exhibits greater uncertainty. Specifically, as reported in Figure \ref{fig:ra_mle_sampler}, the mean activations for semantically plausible agents and patients decrease, along with an increase in the activation of syntactically indicated agents and patients, showing that the model accounts for unusual thematic role assignments in these out-of-distribution (OOD) scenarios, taking into account both semantic and syntactic cues. Refer to Figures \ref{fig:unviolated_sampler} and \ref{fig:violated_sampler} for the mean output activations after each word in the test sentences is input, corresponding to congruent and RA sentences, respectively, using the Bayesian SG model. This behavior highlights the Bayesian model’s enhanced uncertainty to the ambiguity in RA test cases compared to the traditional MLE approach. 

\subsection{Statistical analysis of reversal anomaly sentences}
\label{sec:stat_analysis}
To quantitatively compare the predictive performance of the SG model trained using the MLE approach and the last-layer Bayesian approximation with the dropout deterministic sampler, we conducted a series of statistical tests. As described in Section \ref{sec:exp_conditions}, both model-level and item-level analyses were performed on the RA test sentences. First, a one-sample t-test was conducted to determine whether the mean activations for reversed words (e.g., ‘‘pizza as agent" and ‘‘woman as patient") across 10 runs of the MLE-trained SG model (ADAM) significantly differed from zero (model analysis). Additionally, an item-level analysis was conducted for $n=8$ items. The null hypothesis in both tests was that there is no significant difference between the mean activation and zero. Table \ref{tab:ra_model_analysis} summarizes the p-value results from the t-tests for model analysis.
\begin{table}
    \begin{center}
    \begin{tabular}{@{}lccc@{}}
    \toprule
    Word & Role Probe & ADAM & Dropout deterministic sampler \\ \midrule
    Syntactically indicated agents & Agent & 0.9153 & $< 0.0001$  \\
    Semantically plausible agents & Patient & 0.9029 & $< 0.0001$ \\
    All non-relevant patients & Agent & 0.8691 & 0.6642 \\
    \bottomrule
    \end{tabular}
    \caption{Comparison of p-values from t-tests against zero for SG model output activations on RA test sentences, using ADAM and the dropout deterministic sampler (model analysis). Activations obtained with ADAM show no significant deviation from zero ($p > 0.05$), whereas those from the dropout sampler exhibit a significant difference from zero ($p < 0.05$) across reversed words.}
    \label{tab:ra_model_analysis}
    \end{center}
\end{table}
The results indicate that, for the ADAM-trained SG model, activations across all reversed words do not differ significantly from zero ($p > 0.05$), implying that the mean activations are close to zero relative to sample variability. In contrast, the Bayesian SG model employing the dropout deterministic sampler exhibits significant difference from zero  ($p<0.05$) for reversed words (''pizza as agent" and ''woman as patient"). However, for non-relevant patients (e.g., ''chess as agent"), results indicate no significant difference from zero ($p>0.05$). This additional test for non-relevant patients is meant to demonstrate that the non-zero activations were not due to higher noise, but instead reflected meaningful activations in response to the uncertainty induced by the conflict between semantic and syntactic cues. Results from the item-level t-tests against zero, for both ADAM-trained and the Bayesian SG model with the dropout deterministic sampler are provided in Table \ref{tab:ra_item_analysis}.

Further, we conducted paired t-tests  to compare the activations from the dropout deterministic sampler with those from the ADAM-trained SG model.
\begin{table}
    \begin{center}
    \begin{tabular}{@{}lccc@{}}
    \toprule
    Word & Role Probe & t-statistic & p-value \\ \midrule
    Syntactically indicated agents & Agent &  $4.6858 $ & $< 0.0001$ \\
    Semantically plausible agents & Patient &  $6.1014$ &$< 0.0001$ \\
    All non-relevant patients & Agent &$ -5.4184$ & $< 0.0001$ \\
    \bottomrule
    \end{tabular}
    \caption{Paired t-test comparing output activations from the dropout deterministic sampler and the ADAM-trained SG model on RA test sentences (model analysis). Results show significant differences in mean activations ($p < 0.001$), with positive t-statistics for reversed words indicating greater uncertainty in the Bayesian SG model compared to the ADAM-trained model.}
    \label{tab:paired_model}
    \end{center}
\end{table}
The results of the paired t-tests for model analysis, as summarized in Table~\ref{tab:paired_model}, show a significant difference in mean activations across both reversed words and all non-relevant patients ($p < 0.001$). Positive t-statistics for reversed words indicate that the Bayesian SG model yields higher mean activations than ADAM trained SG model, reflecting greater uncertainty in handling reversal anomalies. Conversely, the negative t-statistics for non-relevant patients suggest that the Bayesian SG model produces lower mean activations for such words being assigned the role of ‘‘agent" compared to the ADAM-trained model. This shows that overall the activations of the Bayesian SG model are more specific. For item analysis, refer to Table \ref{tab:paired_item}.

\section{Discussion}\label{sec:discussion}
In this work, we aim to address a key limitation in current artificial neural network (ANN) models of language comprehension: their inability to represent uncertainty in the presence of ambiguous or conflicting cues, such as those found in reversal anomalies (RAs). Traditional models like the Sentence Gestalt (SG) model, when trained with maximum likelihood estimation (MLE), can exhibit overly deterministic behavior—even in scenarios where human sentence processing reflects significant uncertainty \cite{pmlr-v70-guo17a,Hein_2019_CVPR}. This stands in contrast to extensive findings in cognitive science showing that human comprehension involves flexible interpretation and often displays uncertainty under non-canonical structures \cite{ferreira2003, kuperberg2003}.

In accordance with these observations, our comparison of a standard MLE-trained SG model \cite{STJOHN1990217}, and a Bayesian version of the model using an ensemble Kalman filter (EnKF)-based sampler \cite{bhandari2024}, highlights substantial differences in how these approaches deal with semantic and syntactic conflicts. In semantically congruent sentences, both models produced high mean activations for expected agents and patients. However, under RA conditions, the traditional SG model maintained high confidence in semantically plausible but syntactically incorrect role assignments. This indicates a continued reliance on learned event probabilities, failing to reflect the kind of uncertainty that is typically observed in human sentence processing when syntactic and semantic cues are misaligned.

In contrast, the Bayesian SG model displayed behavior more consistent with human comprehension. Specifically, it demonstrated reduced activations for semantically plausible but syntactically incorrect role assignments and increased activations for syntactically indicated roles in RA conditions. This shift reflects greater sensitivity to cue conflict and suggests that the model is capable of maintaining uncertainty in cases where the correct interpretation is ambiguous—bringing it closer to the variability and probabilistic nature of human responses \cite{Kutas2011, ferreira2003}.

These results strengthen the argument that incorporating Bayesian inference into ANN models offers a promising route toward addressing a longstanding gap in modeling human language comprehension. By framing language understanding as a Bayesian inverse problem and using EnKF-based sampling methods \cite{bhandari2024}, we not only obtain better-calibrated predictions but also a model architecture that naturally reflects the uncertainty observed in human behavior. Notably, the Bayesian SG model differentiates between meaningful and irrelevant cues, maintaining low confidence for implausible role assignments while remaining responsive to both semantic and syntactic inputs.

In summary, our findings suggest that deterministic ANNs, while powerful, by themselves may not be well-suited to capturing the ambiguity inherent in real-world language comprehension. In contrast, the Bayesian approach introduced in this paper provides a compelling extension by allowing model predictions to reflect degrees of belief rather than point estimates, thereby aligning better with human cognitive processes. Future work can build on this foundation by exploring more complex linguistic structures, incorporating richer semantic environments, or extending Bayesian inference across varying layers of the model architecture. This line of research may help bridge the gap between computational modeling and the probabilistic nature of human sentence understanding.

\section*{CRediT authorship contribution statement}
\textbf{Diksha Bhandari:} Writing – original draft, Writing – review \& editing, Visualization, Software, Methodology, Investigation, Formal analysis, Data curation, Conceptualization.
\textbf{Alessandro Lopopolo:} Writing – original draft, Writing – review \& editing, Conceptualization.
\textbf{Milena Rabovsky:} Writing – review \& editing, Visualization, Supervision, Project administration, Funding acquisition, Conceptualization.
\textbf{Sebastian Reich:} Writing – review \& editing, Supervision, Project administration, Funding acquisition, Conceptualization.

\appendix
\section{Words and their semantic representations}
\label{sec:corpus_words}
\begin{longtable}{p{3cm}p{10cm}}
\noalign{\global\arrayrulewidth=1pt}\hline 
\textbf{Words} & \textbf{Semantic representations} \\
\hline 
\noalign{\global\arrayrulewidth=0.4pt}
Woman & person, active, adult, female, woman \\
Man & person, active, adult, male, man \\
Girl & person, active, child, female, girl \\
Boy & person, active, child, male, boy \\
\hline
Drink & action, consume, done with liquids, drink \\
Eat & action, consume, done with foods, eat \\
Feed & action, done to animals, done with food, feed \\
Fish & action, done to fishes, done close to water, fish \\
Plant & action, done to plants, done with earth, plant \\
Water & action, done to plants, done with water, water \\
Play & action, done with games, done for fun, play \\
Wear & action, done with clothes, done for warming, wear \\
Read & action, done with letters, perceptual, read \\
Write & action, done with letters, productive, write \\
Look at & action, visual look at \\
Like & action, positive, like \\
\hline
Kitchen & location, inside, place to eat, kitchen \\
Living room & location, inside, place for leisure, living room \\
Bedroom & location, inside, place to sleep, bedroom \\
Garden & location, outside, place for leisure, garden \\
Lake & location, outside, place with animals, lake \\
Park & location, outside, place with animals, park \\
Balcony & location, outside, place to step out, balcony \\
River & location, outside, place with water, river \\
Backyard & location, outside, place behind house, backyard \\
Veranda & location, outside, place in front of house, veranda \\
\hline
Breakfast & situation, food related, in the morning, breakfast \\
Dinner & situation, food related, in the evening, dinner \\
Excursion & situation, going somewhere, to enjoy, excursion \\
Afternoon & situation, after lunch, day time, afternoon \\
Holiday & situation, special day, no work, holiday \\
Sunday & situation, free time, to relax, Sunday \\
Morning & situation, early, wake up, morning \\
Evening & situation, late, get tired, evening \\
\hline
Egg & consumable, food, white, egg \\
Toast & consumable, food, brown, toast \\
Cereals & consumable, food, healthy, cereals \\
Soup & consumable, food, in bowl, soup \\
Pizza & consumable, food, round, pizza \\
Salad & consumable, food, light, salad \\
\hline
Iced tea & consumable, drink, from leaves, iced tea \\
Juice & consumable, drink, from fruit, juice \\
Lemonade & consumable, drink, sweet, lemonade \\
Cacao & consumable, drink, with chocolate, cacao \\
Tea & consumable, drink, hot, tea \\
Coffee & consumable, drink, activating, coffee \\
\hline
Chess & game, entertaining, strategic, chess \\
Monopoly & game, entertaining, with dice, monopoly \\
Backgammon & game, entertaining, old, backgammon \\
\hline
Jeans & garment, to cover body, for legs, jeans \\
Shirt & garment, to cover body, for upper part, shirt \\
Pajamas & garment, to cover body, for night, pajamas \\
\hline
Novel & contains language, contains letters, art, novel \\
Email & contains language, contains letters, communication, email \\
SMS & contains language, contains letters, communication, short, SMS \\
Letter & contains language, contains letters, communication, on paper, letter \\
Paper & contains language, contains letters, scientific, paper \\
Newspaper & contains language, contains letters, information, newspaper \\
\hline
Rose & can grow, has roots, has petals, red, rose \\
Daisy & can grow, has roots, has petals, yellow, daisy \\
Tulip & can grow, has roots, has petals, colorful, tulip \\
\hline
Pine & can grow, has roots, has bark, green, pine \\
Oak & can grow, has roots, has bark, tall, oak \\
Birch & can grow, has roots, has bark, white bark, birch \\
\hline
Robin & can grow, can move, can fly, red, robin \\
Canary & can grow, can move, can fly, yellow, canary \\
Sparrow & can grow, can move, can fly, brown, sparrow \\
\hline
Sunfish & can grow, can move, can swim, yellow, sunfish \\
Salmon & can grow, can move, can swim, red, salmon \\
Eel & can grow, can move, can swim, long, eel \\
\hline
During/at & no output units (activated together with situation words, e.g., ``at breakfast") \\
In & no output units (activated together with location words, e.g., ``in the park") \\
\noalign{\global\arrayrulewidth=1pt}\hline 
\caption{Words (input unit labels) and their corresponding semantic representations (output unit labels representing the concepts referred to by the words).}
\label{tab:words_corpus}
\end{longtable}

\section{Output activations for congruent and reversal anomaly test sentences}\label{sec:appendix_activations}
\begin{figure}[H]
\parbox{\textwidth}{\textbf{\enquote{During dinner...}}}
    \centering
    \begin{subfigure}[b]{0.243\textwidth}
        \centering
        \includegraphics[width=\textwidth]{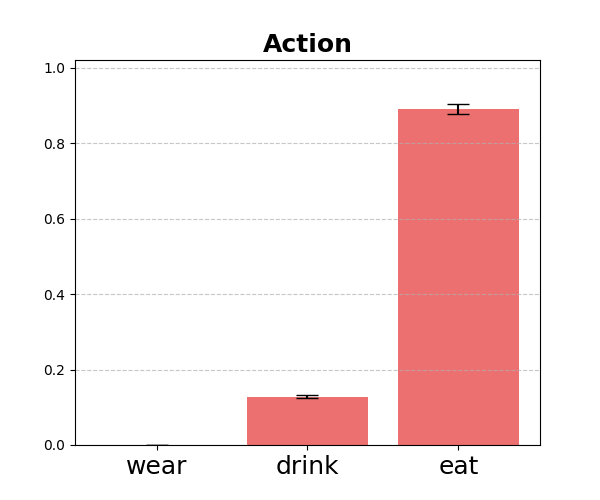}
    \end{subfigure}
    \begin{subfigure}[b]{0.243\textwidth}
        \centering
        \includegraphics[width=\textwidth]{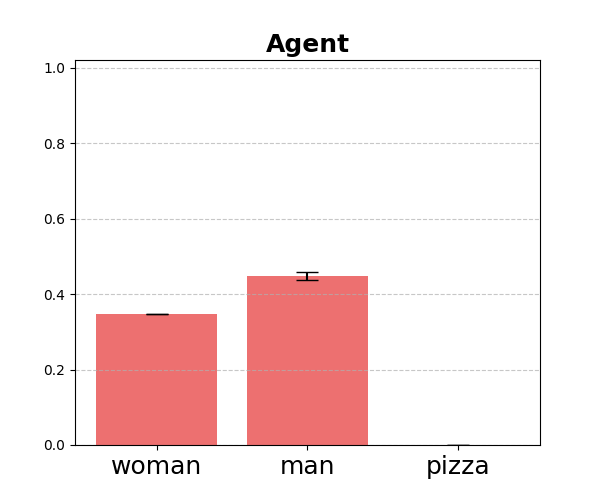}
    \end{subfigure}
    \begin{subfigure}[b]{0.243\textwidth}
        \centering
        \includegraphics[width=\textwidth]{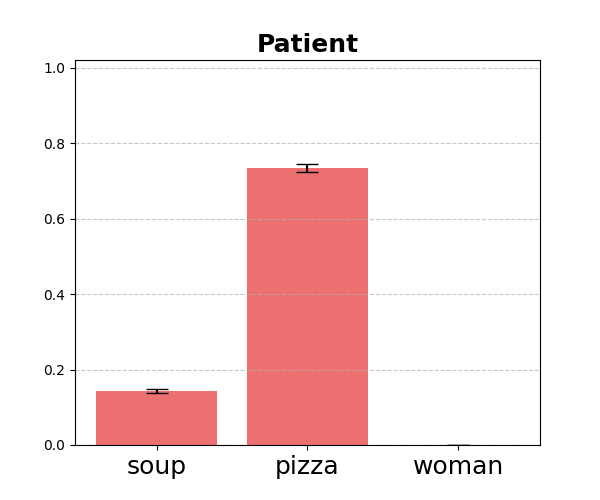}
    \end{subfigure}
    \begin{subfigure}[b]{0.243\textwidth}
        \centering
        \includegraphics[width=\textwidth]{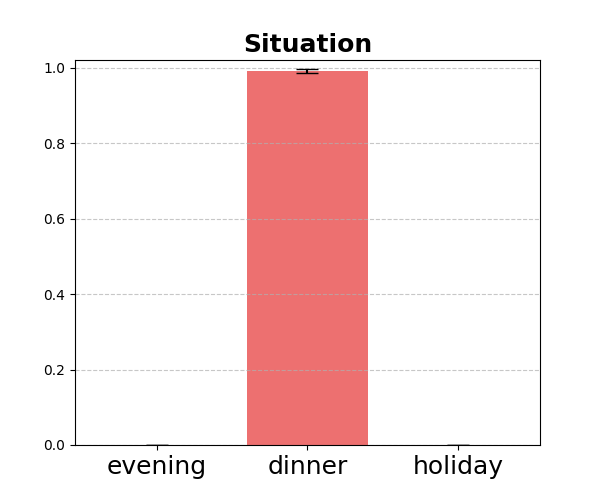}
    \end{subfigure}
    \vspace{0.8em}
\parbox{\textwidth}{\textbf{\enquote{...woman...}}}
    \centering
    \begin{subfigure}[b]{0.243\textwidth}
        \centering
        \includegraphics[width=\textwidth]{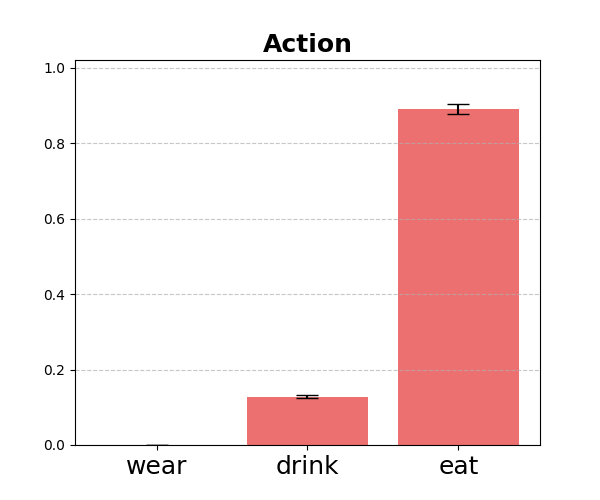}
    \end{subfigure}
    \begin{subfigure}[b]{0.243\textwidth}
        \centering
        \includegraphics[width=\textwidth]{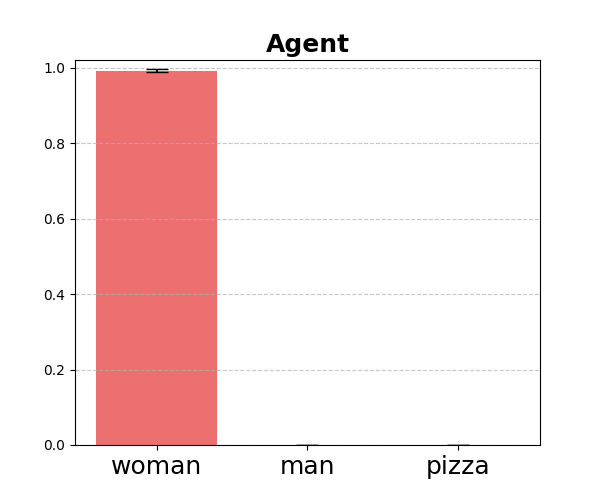}
    \end{subfigure}
    \begin{subfigure}[b]{0.243\textwidth}
        \centering
        \includegraphics[width=\textwidth]{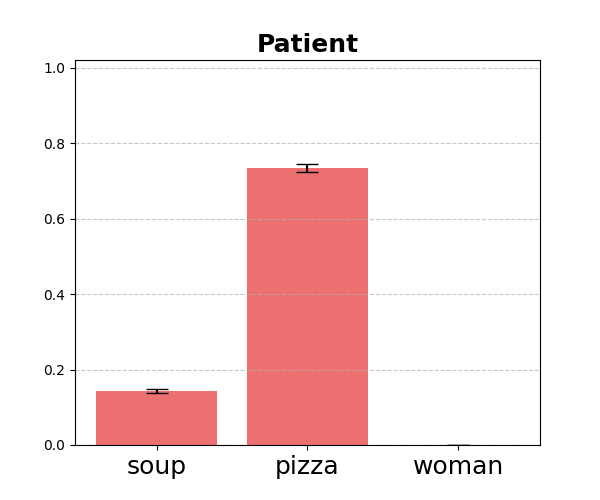}
    \end{subfigure}
    \begin{subfigure}[b]{0.243\textwidth}
        \centering
        \includegraphics[width=\textwidth]{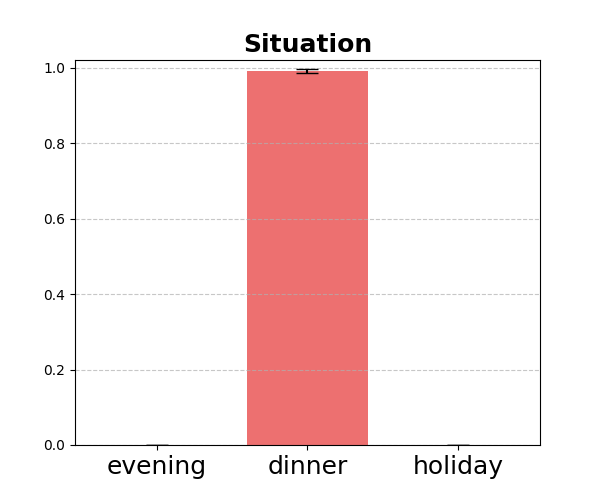}
    \end{subfigure}
     \vspace{0.8em}
\parbox{\textwidth}{\textbf{\enquote{...eats...}}}
    \centering
    \begin{subfigure}[b]{0.243\textwidth}
        \centering
        \includegraphics[width=\textwidth]{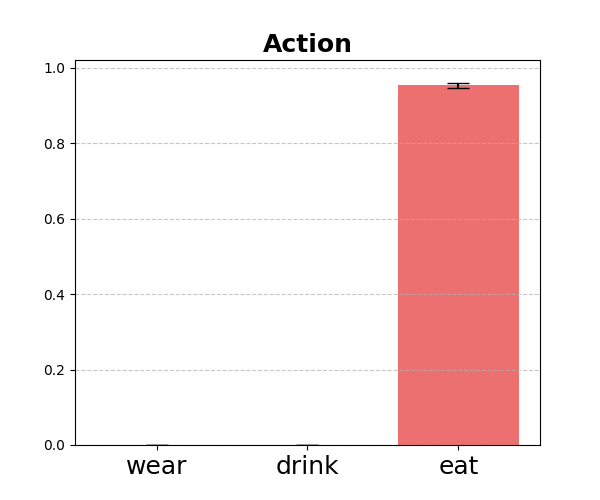}
    \end{subfigure}
    \begin{subfigure}[b]{0.243\textwidth}
        \centering
        \includegraphics[width=\textwidth]{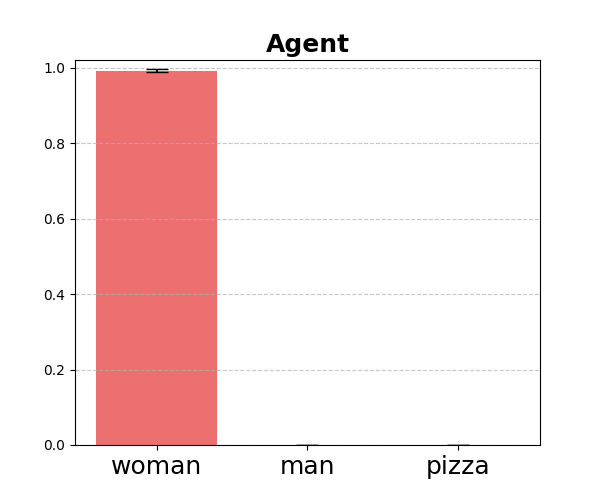}
    \end{subfigure}
    \begin{subfigure}[b]{0.243\textwidth}
        \centering
        \includegraphics[width=\textwidth]{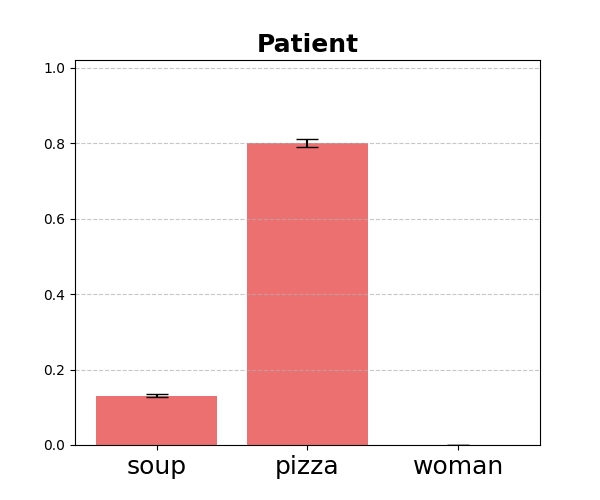}
    \end{subfigure}
    \begin{subfigure}[b]{0.243\textwidth}
        \centering
        \includegraphics[width=\textwidth]{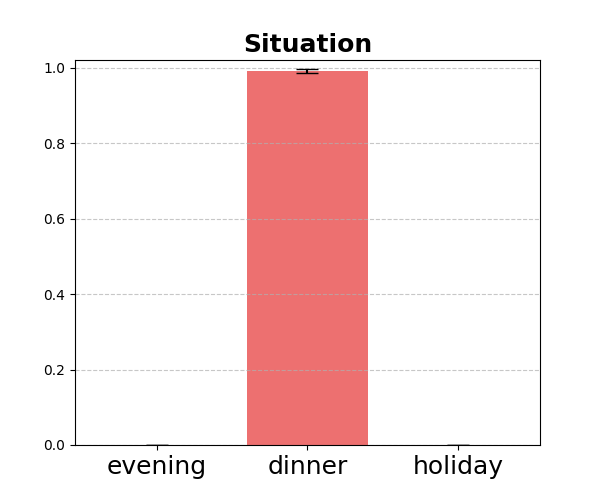}
    \end{subfigure}
     \vspace{0.8em}
\parbox{\textwidth}{\textbf{\enquote{...pizza}}}
    \centering
    \begin{subfigure}[b]{0.243\textwidth}
        \centering
        \includegraphics[width=\textwidth]{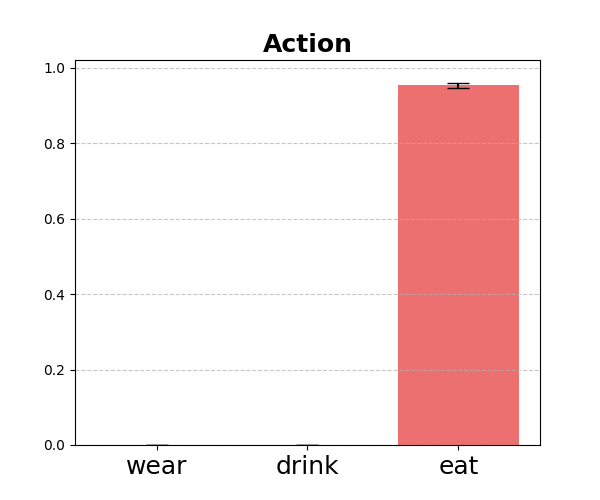}
    \end{subfigure}
    \begin{subfigure}[b]{0.243\textwidth}
        \centering
        \includegraphics[width=\textwidth]{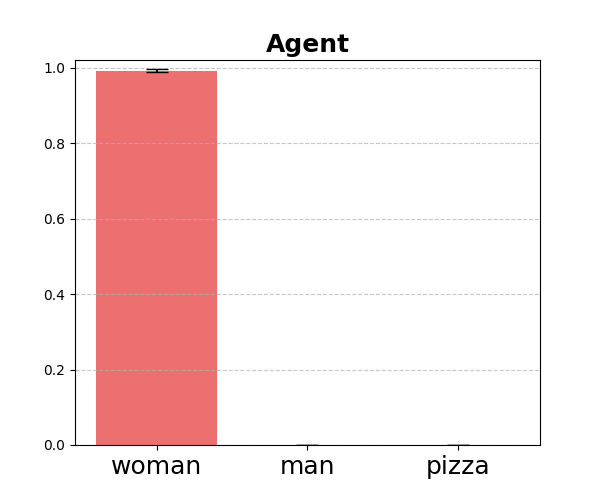}
    \end{subfigure}
    \begin{subfigure}[b]{0.243\textwidth}
        \centering
        \includegraphics[width=\textwidth]{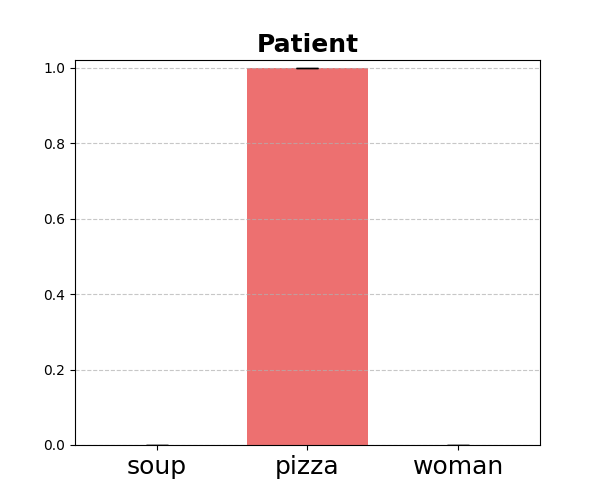}
    \end{subfigure}
    \begin{subfigure}[b]{0.243\textwidth}
        \centering
        \includegraphics[width=\textwidth]{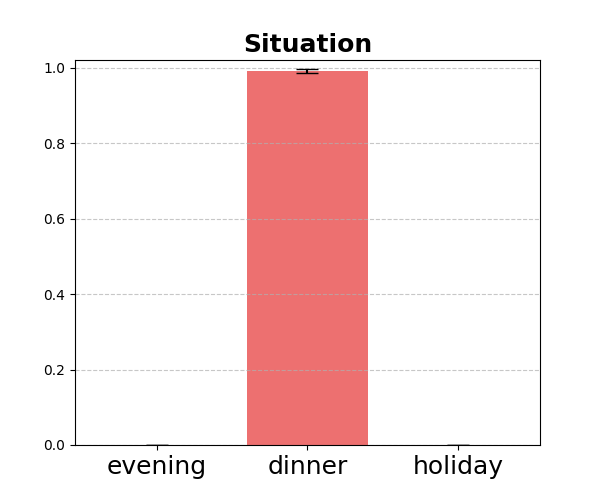}
    \end{subfigure}
    \caption{ADAM trained SGM: mean activations (with standard deviations) of selected output units for congruent form of test sentences.}
    \label{fig:unviolated_mle}
\end{figure}

\begin{figure}[H]
\parbox{\textwidth}{\textbf{\enquote{During dinner...}}}
    \centering
    \begin{subfigure}[b]{0.243\textwidth}
        \centering
        \includegraphics[width=\textwidth]{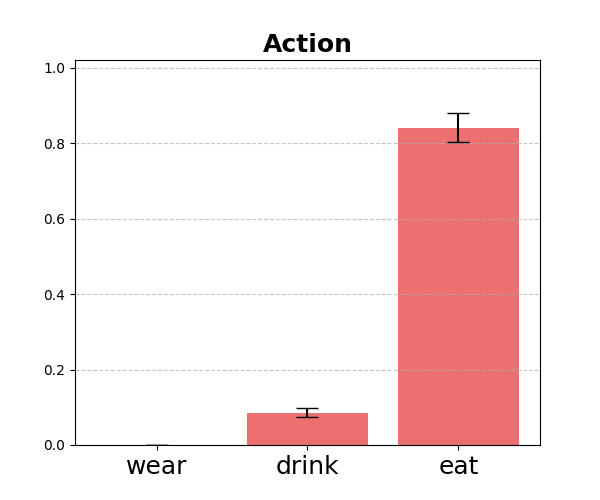}
    \end{subfigure}
    \begin{subfigure}[b]{0.243\textwidth}
        \centering
        \includegraphics[width=\textwidth]{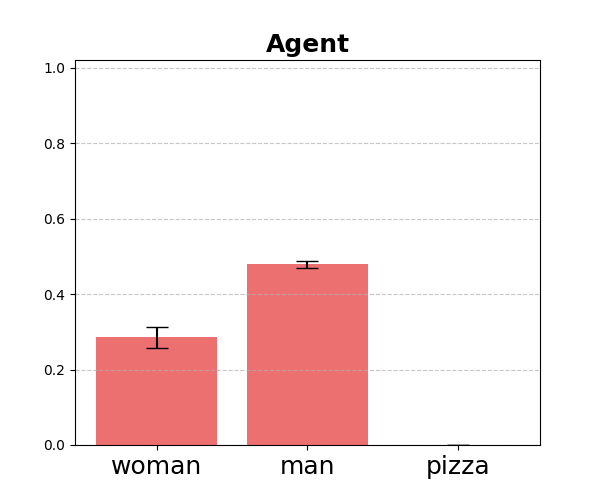}
    \end{subfigure}
    \begin{subfigure}[b]{0.243\textwidth}
        \centering
        \includegraphics[width=\textwidth]{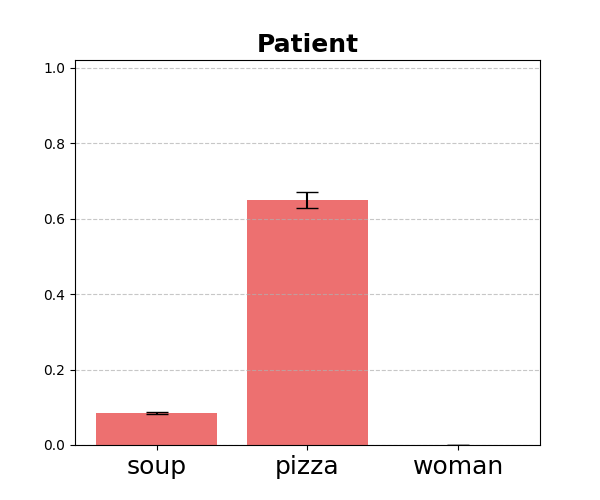}
    \end{subfigure}
    \begin{subfigure}[b]{0.243\textwidth}
        \centering
        \includegraphics[width=\textwidth]{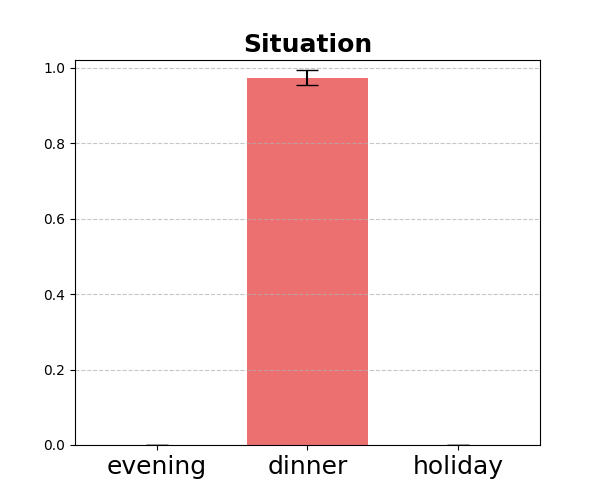}
    \end{subfigure}
    \vspace{0.8em}
\parbox{\textwidth}{\textbf{\enquote{...woman...}}}
    \centering
    \begin{subfigure}[b]{0.243\textwidth}
        \centering
        \includegraphics[width=\textwidth]{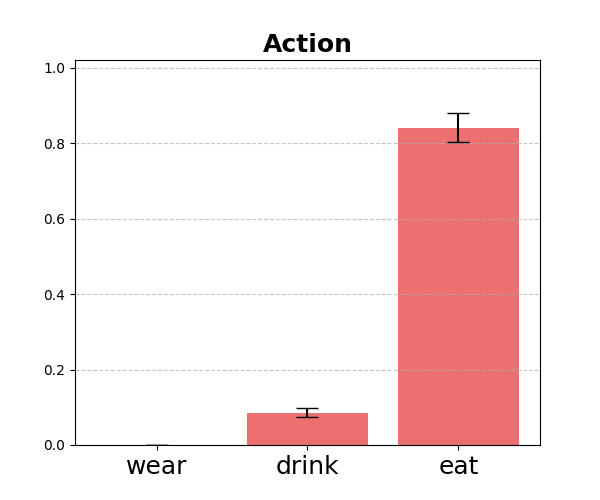}
    \end{subfigure}
    \begin{subfigure}[b]{0.243\textwidth}
        \centering
        \includegraphics[width=\textwidth]{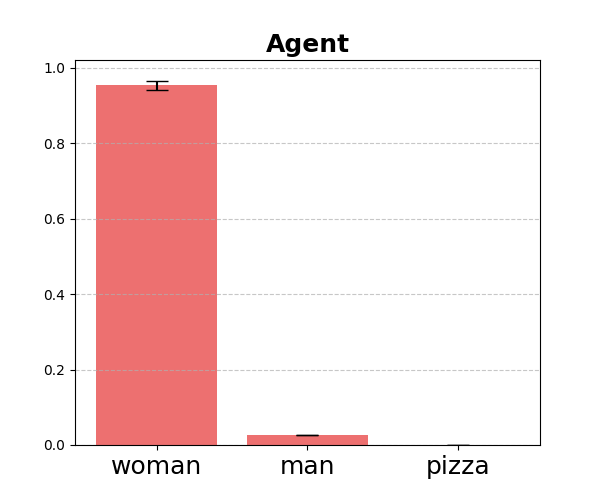}
    \end{subfigure}
    \begin{subfigure}[b]{0.243\textwidth}
        \centering
        \includegraphics[width=\textwidth]{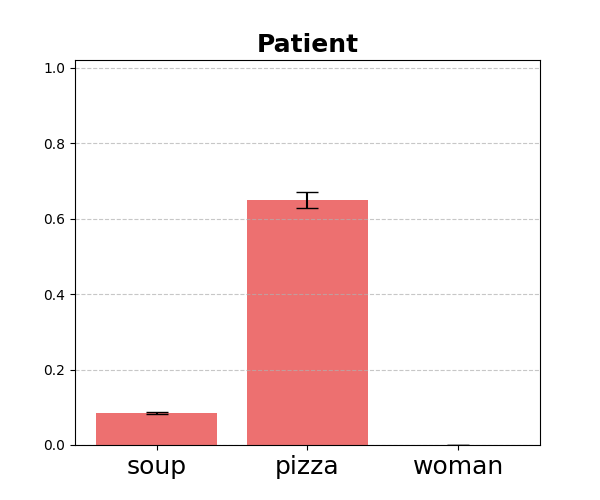}
    \end{subfigure}
    \begin{subfigure}[b]{0.243\textwidth}
        \centering
        \includegraphics[width=\textwidth]{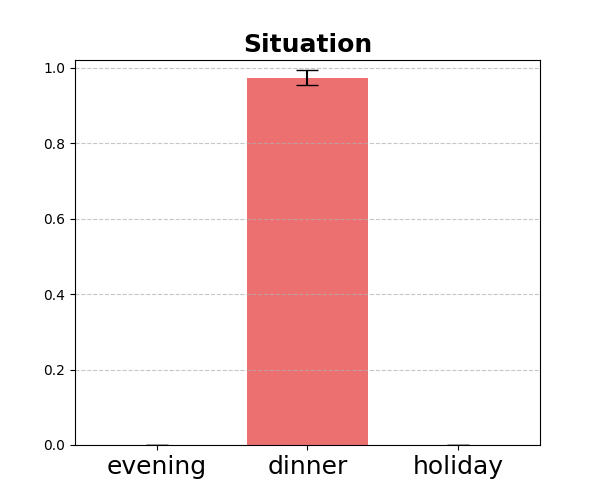}
    \end{subfigure}
     \vspace{0.8em}
\parbox{\textwidth}{\textbf{\enquote{...eats...}}}
    \centering
    \begin{subfigure}[b]{0.243\textwidth}
        \centering
        \includegraphics[width=\textwidth]{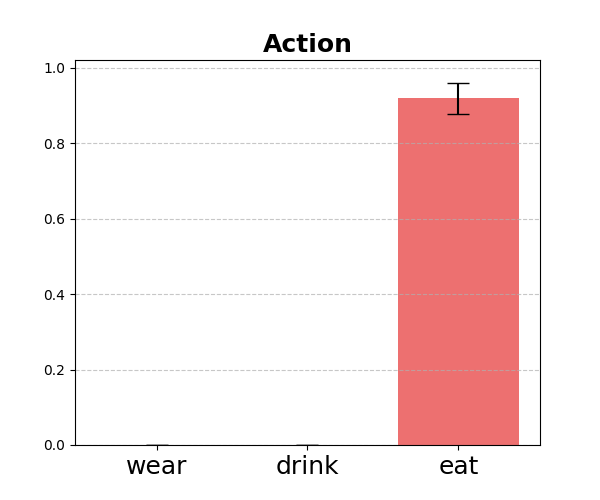}
    \end{subfigure}
    \begin{subfigure}[b]{0.243\textwidth}
        \centering
        \includegraphics[width=\textwidth]{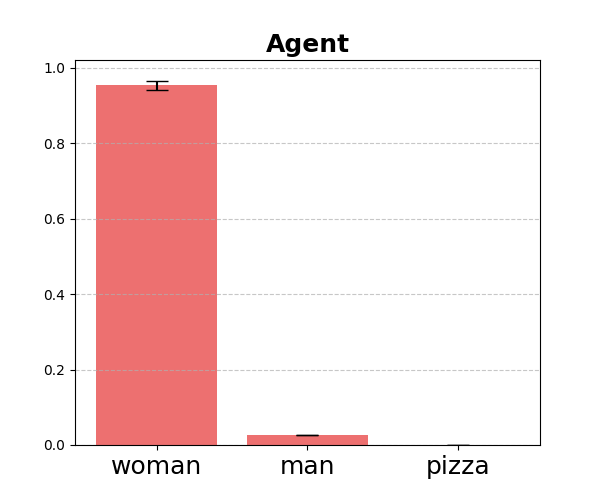}
    \end{subfigure}
    \begin{subfigure}[b]{0.243\textwidth}
        \centering
        \includegraphics[width=\textwidth]{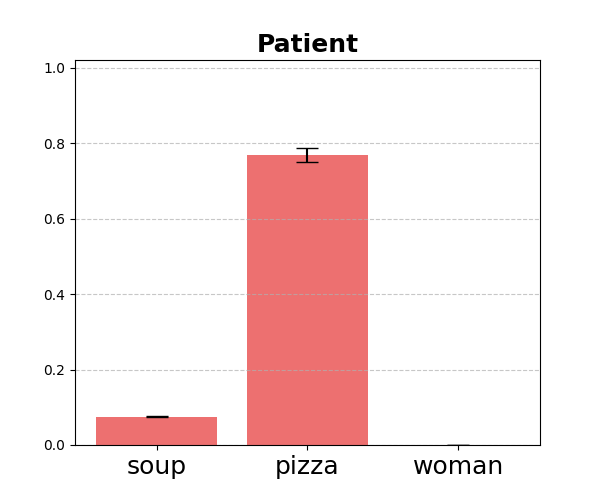}
    \end{subfigure}
    \begin{subfigure}[b]{0.243\textwidth}
        \centering
        \includegraphics[width=\textwidth]{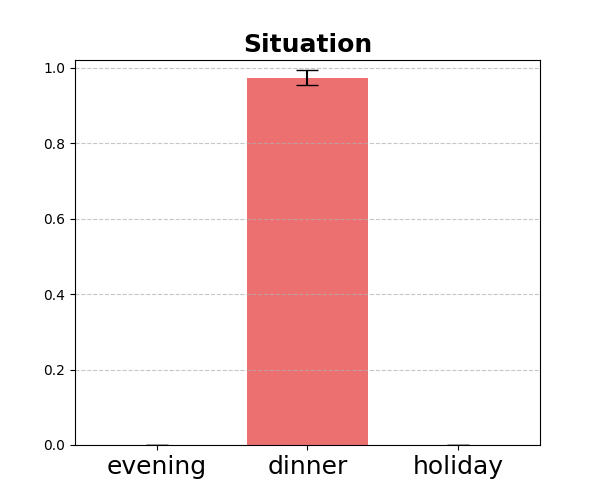}
    \end{subfigure}
     \vspace{0.8em}
\parbox{\textwidth}{\textbf{\enquote{...pizza}}}
    \centering
    \begin{subfigure}[b]{0.243\textwidth}
        \centering
        \includegraphics[width=\textwidth]{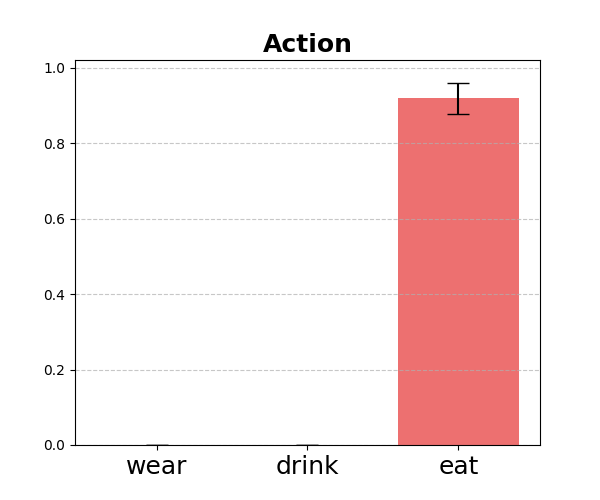}
    \end{subfigure}
    \begin{subfigure}[b]{0.243\textwidth}
        \centering
        \includegraphics[width=\textwidth]{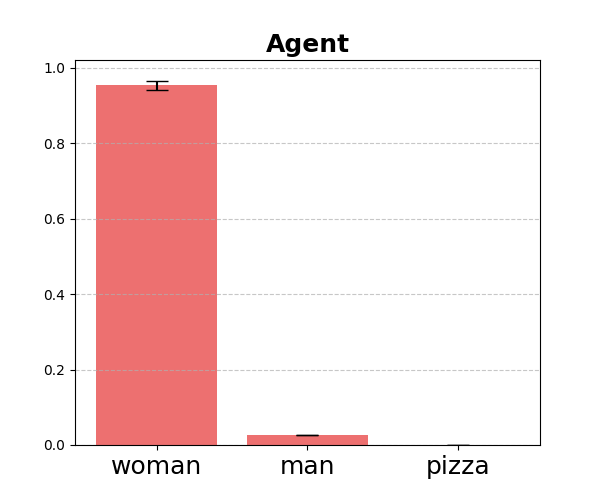}
    \end{subfigure}
    \begin{subfigure}[b]{0.243\textwidth}
        \centering
        \includegraphics[width=\textwidth]{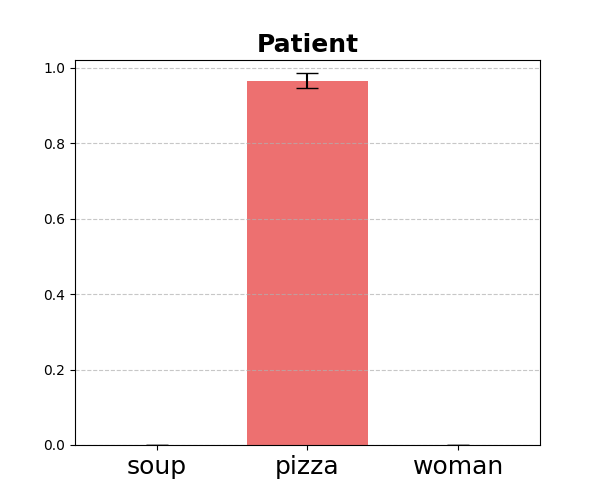}
    \end{subfigure}
    \begin{subfigure}[b]{0.243\textwidth}
        \centering
        \includegraphics[width=\textwidth]{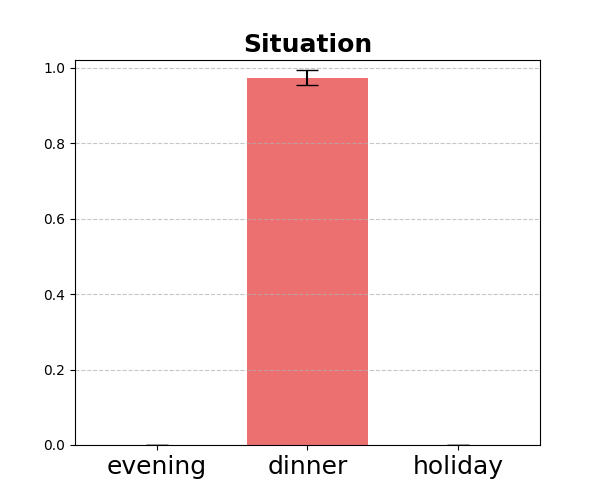}
    \end{subfigure}
    \caption{Dropout deterministic sampler for SGM: mean activations (with standard deviations) of selected output units for congruent form of test sentences.}
    \label{fig:unviolated_sampler}
\end{figure}

\begin{figure}[H]
\parbox{\textwidth}{\textbf{\enquote{During dinner...}}}
    \centering
    \begin{subfigure}[b]{0.243\textwidth}
        \centering
        \includegraphics[width=\textwidth]{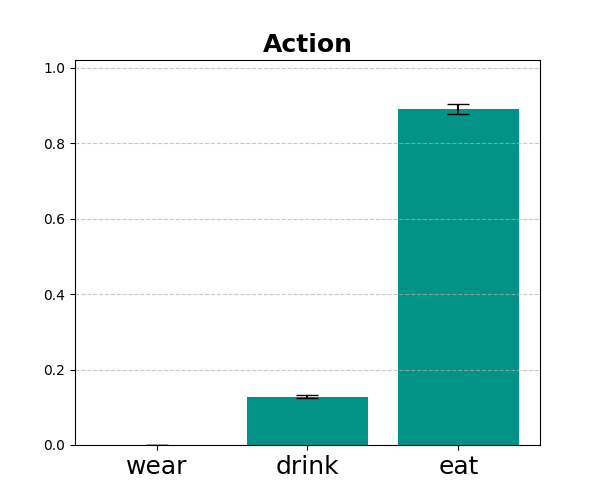}
    \end{subfigure}
    \begin{subfigure}[b]{0.243\textwidth}
        \centering
        \includegraphics[width=\textwidth]{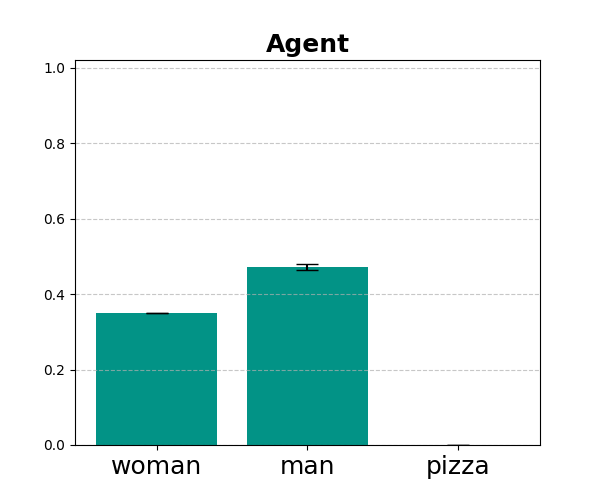}
    \end{subfigure}
    \begin{subfigure}[b]{0.243\textwidth}
        \centering
        \includegraphics[width=\textwidth]{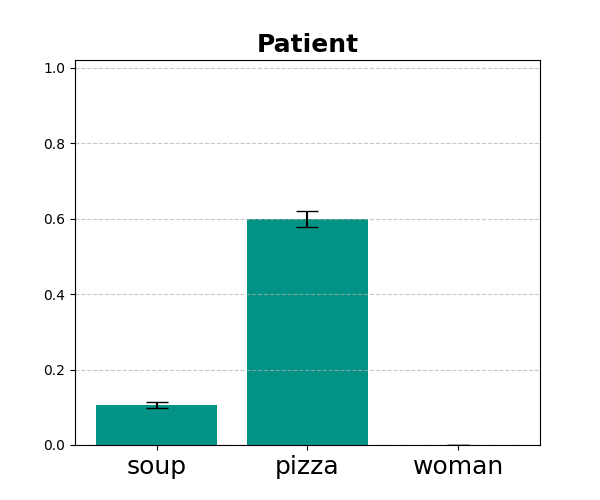}
    \end{subfigure}
    \begin{subfigure}[b]{0.243\textwidth}
        \centering
        \includegraphics[width=\textwidth]{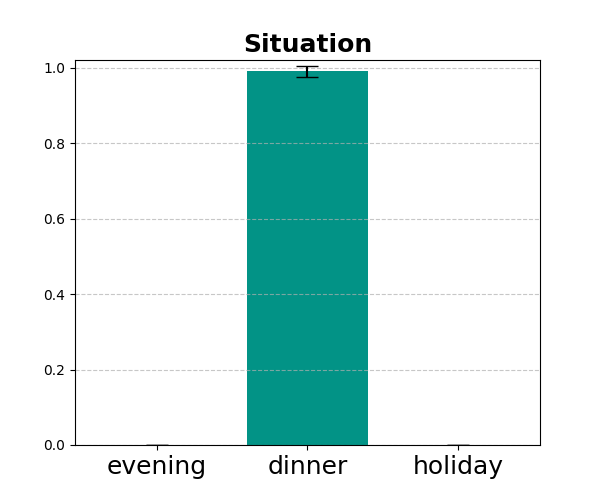}
    \end{subfigure}
    \vspace{0.8em}
\parbox{\textwidth}{\textbf{\enquote{...pizza...}}}
    \centering
    \begin{subfigure}[b]{0.243\textwidth}
        \centering
        \includegraphics[width=\textwidth]{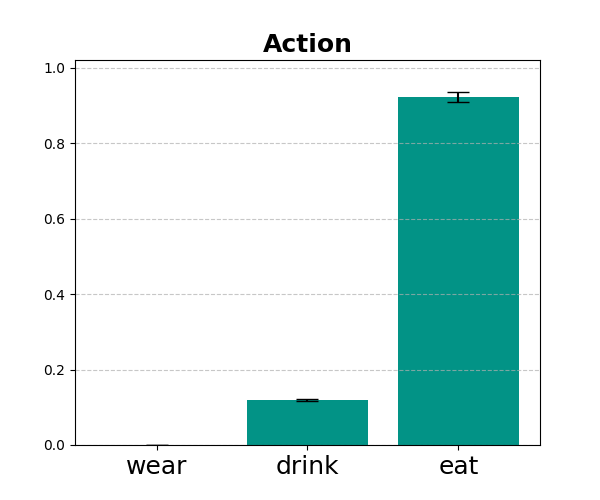}
    \end{subfigure}
    \begin{subfigure}[b]{0.243\textwidth}
        \centering
        \includegraphics[width=\textwidth]{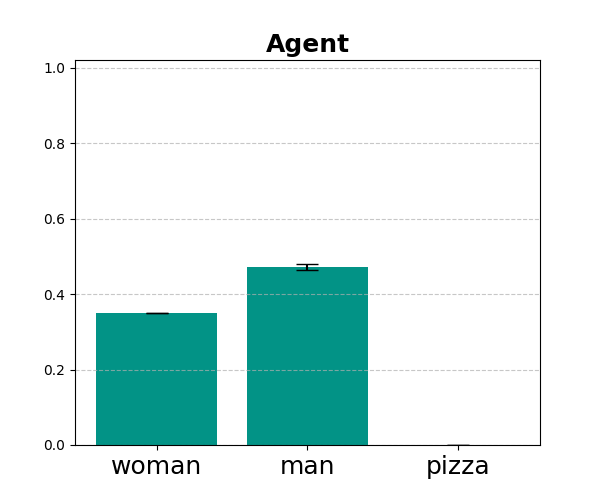}
    \end{subfigure}
    \begin{subfigure}[b]{0.243\textwidth}
        \centering
        \includegraphics[width=\textwidth]{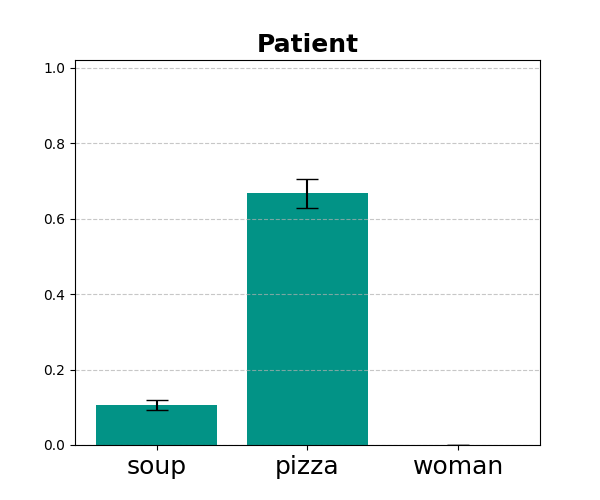}
    \end{subfigure}
    \begin{subfigure}[b]{0.243\textwidth}
        \centering
        \includegraphics[width=\textwidth]{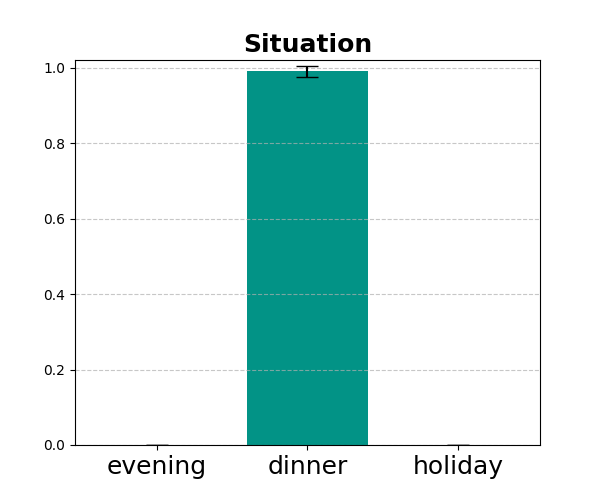}
    \end{subfigure}
     \vspace{0.8em}
\parbox{\textwidth}{\textbf{\enquote{...eats...}}}
    \centering
    \begin{subfigure}[b]{0.243\textwidth}
        \centering
        \includegraphics[width=\textwidth]{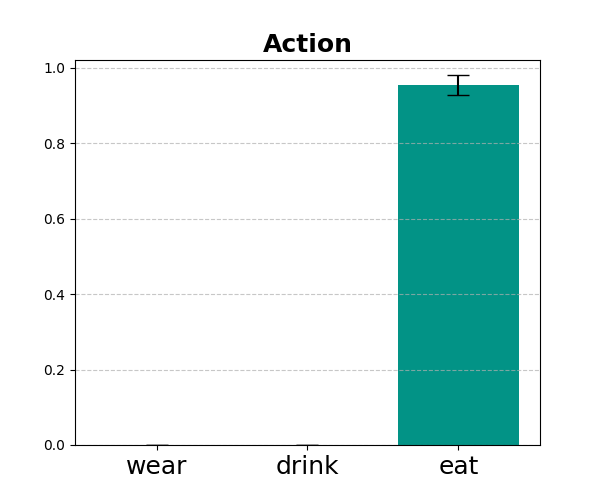}
    \end{subfigure}
    \begin{subfigure}[b]{0.243\textwidth}
        \centering
        \includegraphics[width=\textwidth]{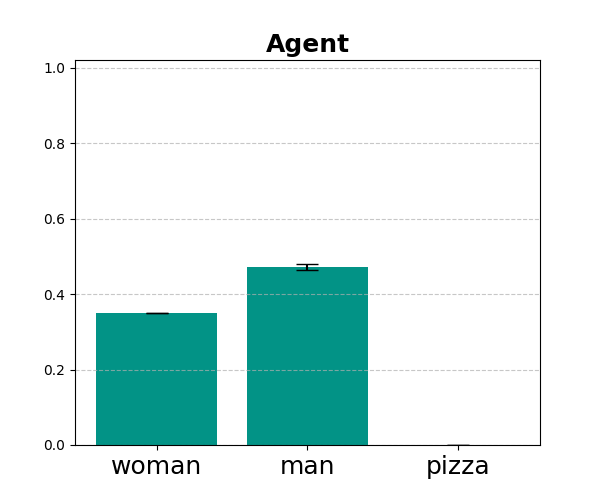}
    \end{subfigure}
     \begin{subfigure}[b]{0.243\textwidth}
        \centering
        \includegraphics[width=\textwidth]{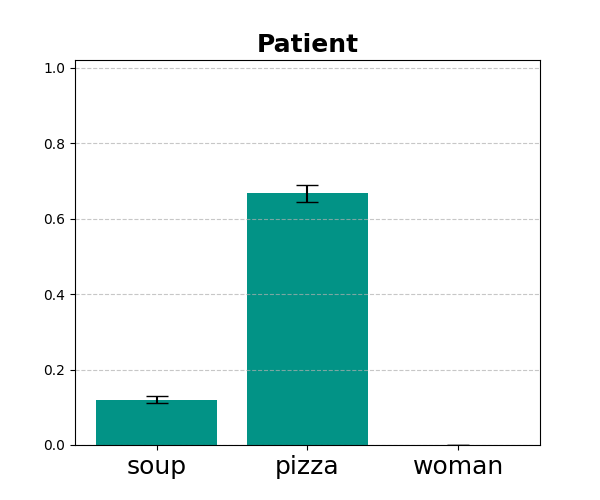}
    \end{subfigure}
    \begin{subfigure}[b]{0.243\textwidth}
        \centering
        \includegraphics[width=\textwidth]{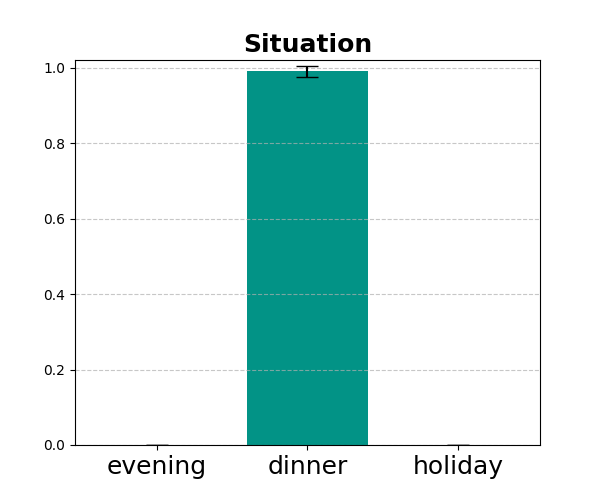}
   \end{subfigure}
    \vspace{0.8em}
\parbox{\textwidth}{\textbf{\enquote{...woman}}}
    \centering
    \begin{subfigure}[b]{0.243\textwidth}
        \centering
        \includegraphics[width=\textwidth]{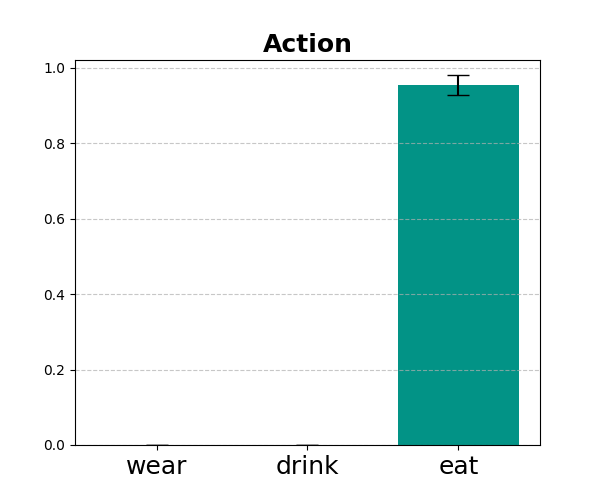}
    \end{subfigure}
    \begin{subfigure}[b]{0.243\textwidth}
        \centering
        \includegraphics[width=\textwidth]{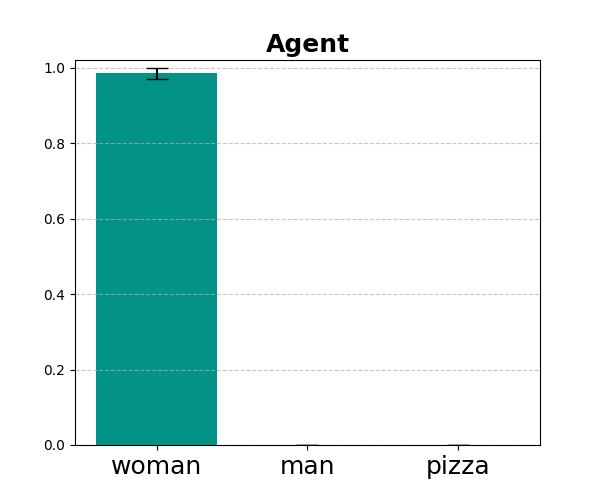}
    \end{subfigure}
    \begin{subfigure}[b]{0.243\textwidth}
        \centering
        \includegraphics[width=\textwidth]{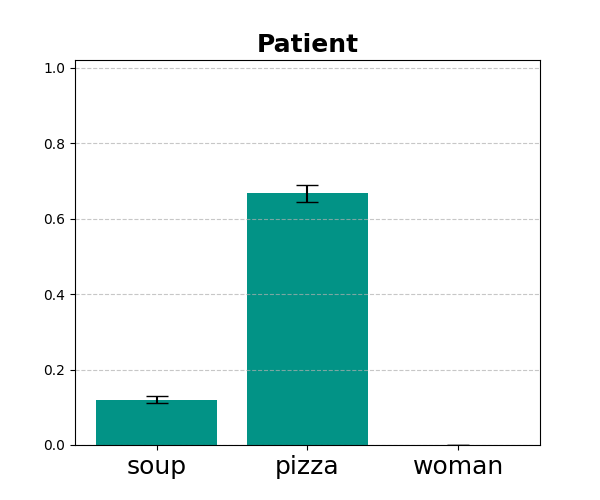}
    \end{subfigure}
    \begin{subfigure}[b]{0.243\textwidth}
        \centering
        \includegraphics[width=\textwidth]{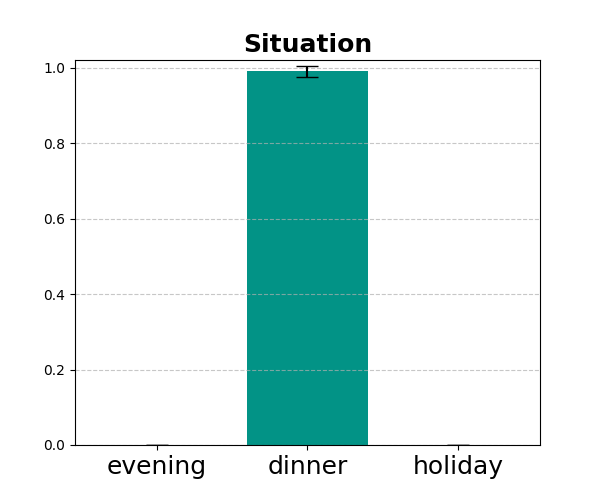}
    \end{subfigure}
    \caption{ADAM trained SGM: mean activations (with standard deviations) of selected output units for RA form of test sentences. Note that the model continues to represent the pizza as the patient (and not the agent) of eating, even after the word “eat” has been presented as input in the sentence.}
    \label{fig:violated_mle}
\end{figure}

\begin{figure}[H]
\parbox{\textwidth}{\textbf{\enquote{During dinner...}}}
    \centering
    \begin{subfigure}[b]{0.243\textwidth}
        \centering
        \includegraphics[width=\textwidth]{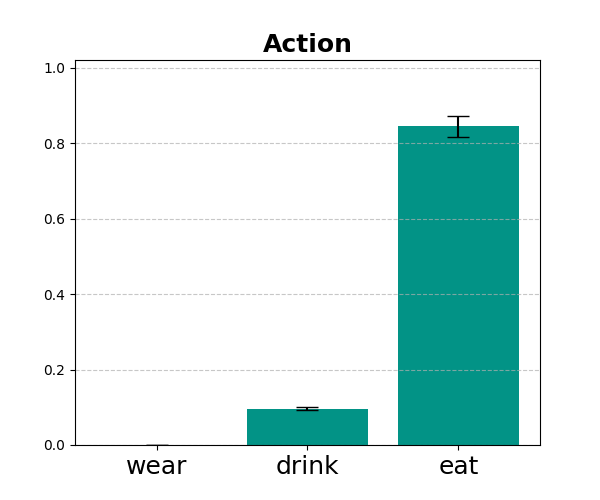}
    \end{subfigure}
    \begin{subfigure}[b]{0.243\textwidth}
        \centering
        \includegraphics[width=\textwidth]{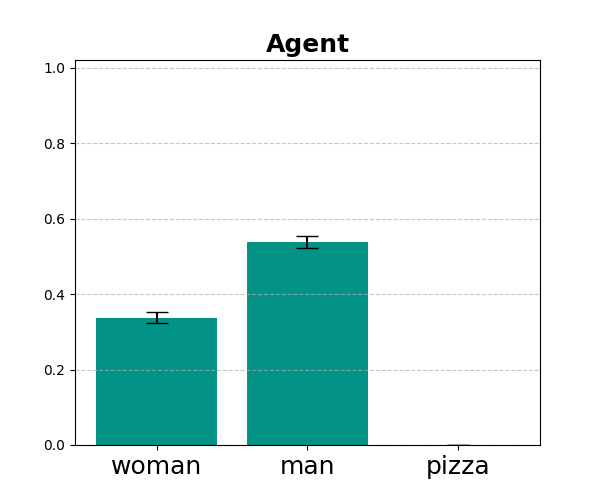}
    \end{subfigure}
    \begin{subfigure}[b]{0.243\textwidth}
        \centering
        \includegraphics[width=\textwidth]{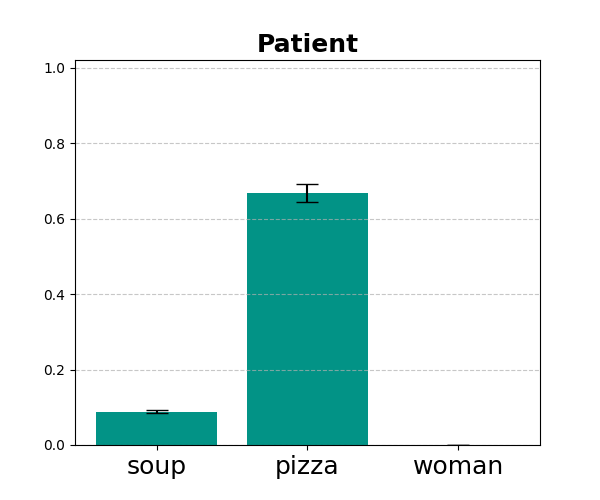}
    \end{subfigure}
    \begin{subfigure}[b]{0.243\textwidth}
        \centering
        \includegraphics[width=\textwidth]{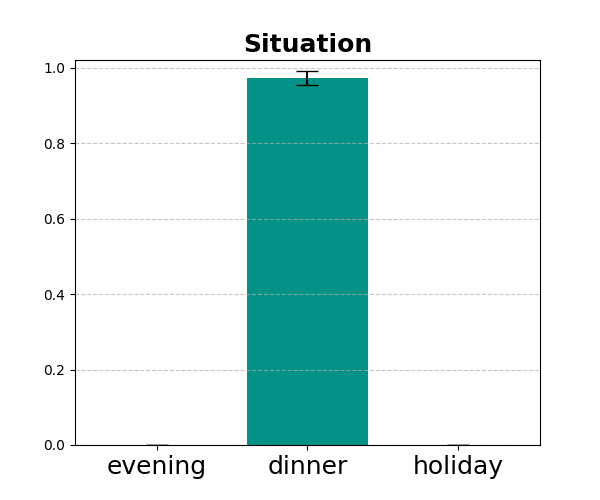}
    \end{subfigure}
    \vspace{0.8em}
\parbox{\textwidth}{\textbf{\enquote{...pizza...}}}
    \centering
    \begin{subfigure}[b]{0.243\textwidth}
        \centering
        \includegraphics[width=\textwidth]{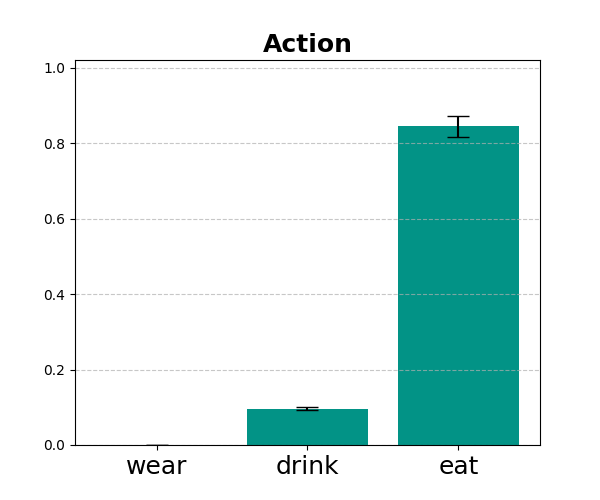}
    \end{subfigure}
    \begin{subfigure}[b]{0.243\textwidth}
        \centering
        \includegraphics[width=\textwidth]{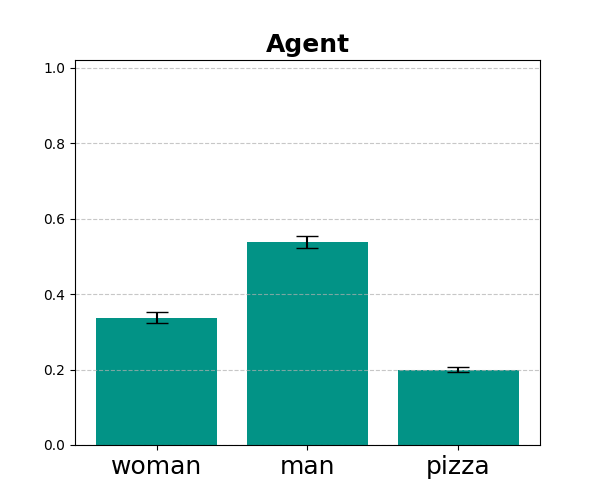}
    \end{subfigure}
    \begin{subfigure}[b]{0.243\textwidth}
        \centering
        \includegraphics[width=\textwidth]{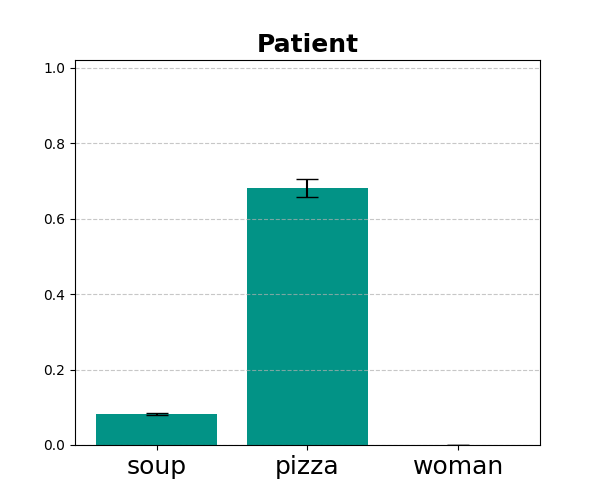}
    \end{subfigure}
    \begin{subfigure}[b]{0.243\textwidth}
        \centering
        \includegraphics[width=\textwidth]{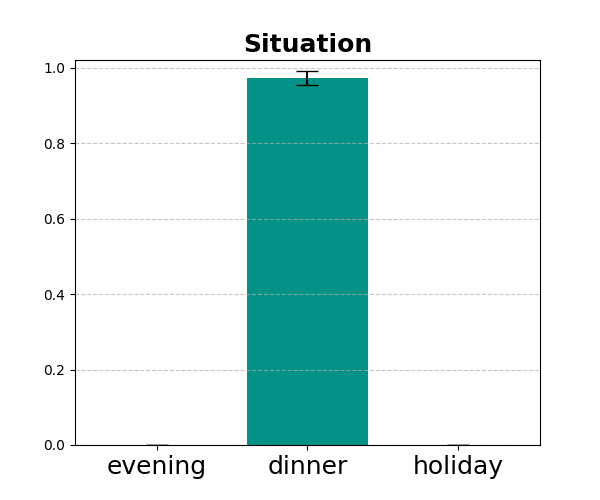}
    \end{subfigure}
     \vspace{0.8em}
\parbox{\textwidth}{\textbf{\enquote{...eats...}}}
    \centering
    \begin{subfigure}[b]{0.243\textwidth}
        \centering
        \includegraphics[width=\textwidth]{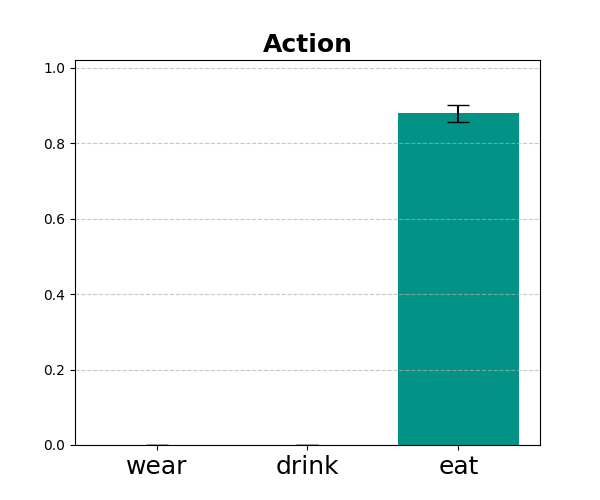}
    \end{subfigure}
    \begin{subfigure}[b]{0.243\textwidth}
        \centering
        \includegraphics[width=\textwidth]{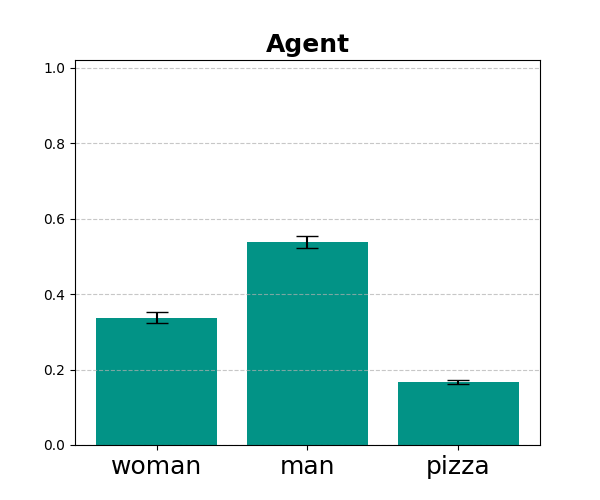}
    \end{subfigure}
     \begin{subfigure}[b]{0.243\textwidth}
        \centering
        \includegraphics[width=\textwidth]{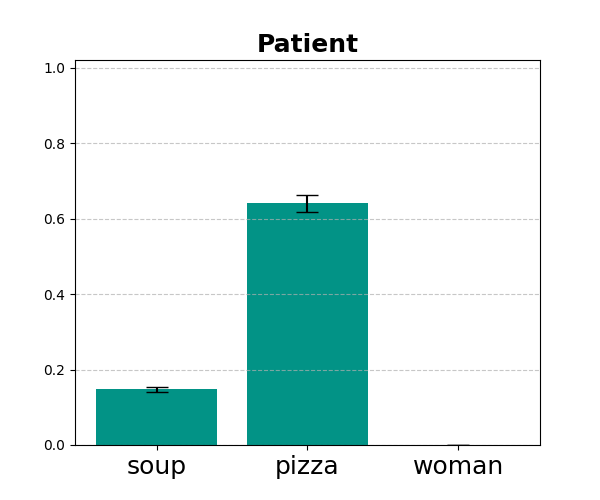}
    \end{subfigure}
    \begin{subfigure}[b]{0.243\textwidth}
        \centering
        \includegraphics[width=\textwidth]{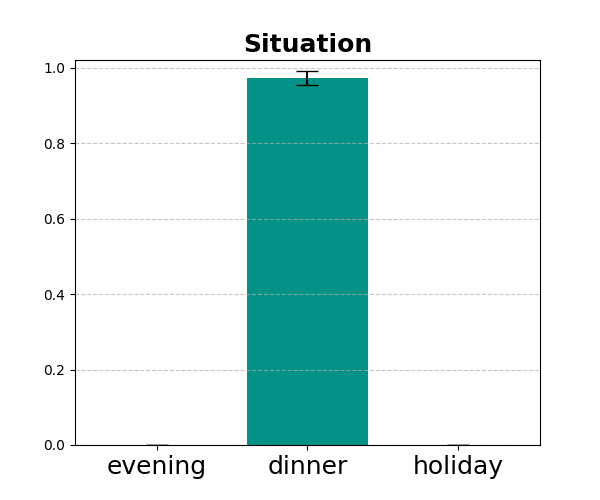}
   \end{subfigure}
    \vspace{0.8em}
\parbox{\textwidth}{\textbf{\enquote{...woman}}}
    \centering
    \begin{subfigure}[b]{0.243\textwidth}
        \centering
        \includegraphics[width=\textwidth]{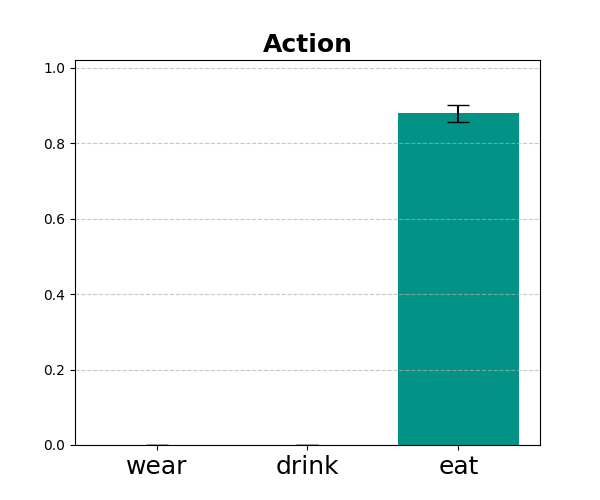}
    \end{subfigure}
    \begin{subfigure}[b]{0.243\textwidth}
        \centering
        \includegraphics[width=\textwidth]{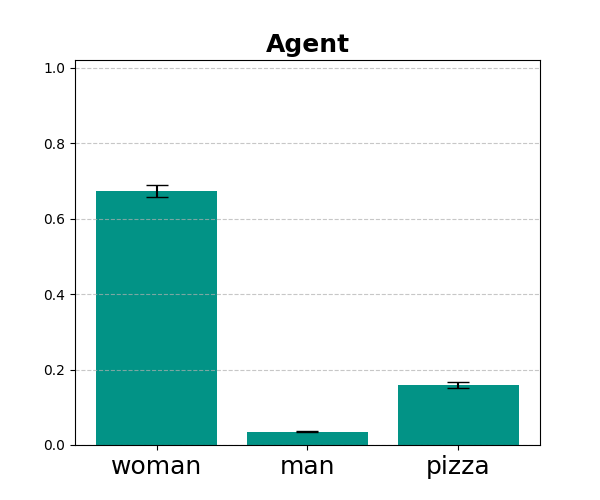}
    \end{subfigure}
    \begin{subfigure}[b]{0.243\textwidth}
        \centering
        \includegraphics[width=\textwidth]{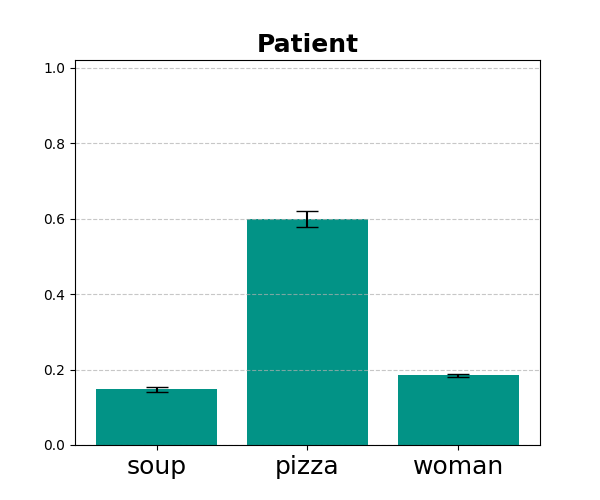}
    \end{subfigure}
    \begin{subfigure}[b]{0.243\textwidth}
        \centering
        \includegraphics[width=\textwidth]{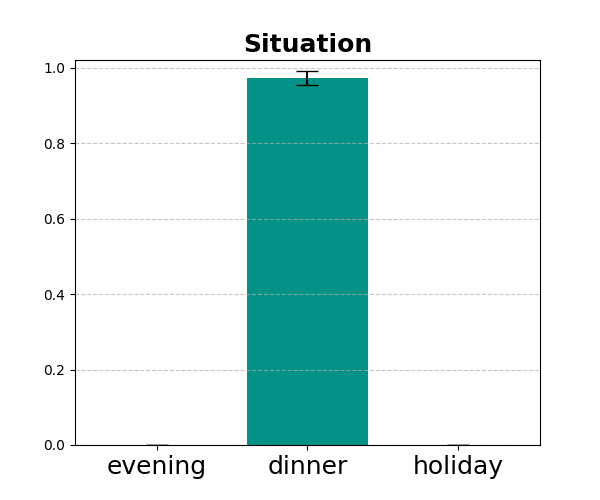}
    \end{subfigure}
    \caption{Dropout deterministic sampler for SGM: mean activations (with standard deviations) of selected output units for RA form of test sentences. Note that the model shows uncertainty while representing pizza (and woman) as the agent (and the patient) of eating, after the word “eat” has been presented as input in the sentence.}
    \label{fig:violated_sampler}
\end{figure}

\section{Statistical (item) analysis of reversal anomaly sentences}\label{sec:appendix_ttests}
Table \ref{tab:ra_item_analysis} summarizes the results of the one-sample t-tests against zero, conducted for item analysis ($n=8$ items). 
\begin{table}[H]
    \begin{center}
    \begin{tabular}{@{}lccc@{}}
    \toprule
    Word & Role Probe & ADAM & Dropout deterministic sampler \\ \midrule
    Syntactically indicated agents & Agent & 0.8890 & $< 0.0001$  \\
    Semantically plausible agents & Patient & 0.8606 & $< 0.0001$ \\
    All non-relevant patients & Agent & 0.8906 & 0.7140  \\
    \bottomrule
    \end{tabular}
    \caption{Comparison of p-values from t-tests against zero for SGM output activations obtained using ADAM and the dropout deterministic sampler on RA test sentences (item analysis). The results show no significant deviation from zero for ADAM ($p > .05$), while the dropout sampler shows significant differences from zero for activations across reversed words ($p < .05$).}
    \label{tab:ra_item_analysis}
    \end{center}
\end{table}
Table \ref{tab:paired_item} presents the results of paired t-tests comparing the activations generated by the Bayesian SG model and the ADAM-trained SG model ($n=8$ items).
\begin{table}[H]
    \begin{center}
    \begin{tabular}{@{}lccc@{}}
    \toprule
    Word & Role Probe & t-statistic & p-value \\ \midrule
    Syntactically indicated agents & Agent &  $9.1730$ & $< 0.0001$ \\
    Semantically plausible agents & Patient &  $13.140$ &$< 0.0001$ \\
    All non-relevant patients & Agent & $-9.8349$ & $< 0.0001$ \\
    \bottomrule
    \end{tabular}
   \caption{Paired t-test comparing activations from the dropout deterministic sampler to the ADAM-trained SGM for RA test sentences (item analysis). The results show significant differences in mean activations ($p < .001$), with positive t-statistics for reversed words indicating higher uncertainty in the Bayesian SGM compared to the ADAM-trained model.}
    \label{tab:paired_item}
    \end{center}
\end{table}

\section{Effect of modifying prior covariance \texorpdfstring{$P_{\text{prior}}$}{Pprior}}\label{sec_appendix_prior}
We investigate the impact of varying the prior covariance \( P_{\text{prior}} \) on model activations, focusing on the behavior of the dropout deterministic sampler in RA test sentences. The experiment evaluates model activations for three different prior covariance $P_{\text{prior}}$: \( 0.01 \operatorname{I} \), \( \operatorname{I} \), and \( 5 \operatorname{I} \), where \( \operatorname{I} \) is the identity matrix (scaled by different magnitudes).
\begin{table}[H]
    \begin{center}
    \begin{tabular}{@{}l  c  ccc@{}}
    \toprule
    Word &   Role Probe & \multicolumn{3}{c}{p-value} \\ 
    \cmidrule(lr){3-5} 
    & & $0.01 \operatorname{I}$ & $\operatorname{I}$ & $5 \operatorname{I}$  \\ \midrule
    Syntactically indicated agents & Agent & $0.2747$ & $<0.001$ & $<0.001$ \\
    Semantically plausible agents & Patient & $0.2110$ & $<0.001$ & $<0.001$ \\
    All non-relevant patients & Agent  & $0.9635$ & $<0.001$ & $<0.001$ \\ 
    \bottomrule
    \end{tabular}
   \caption{Paired t-test comparing activations from the ADAM-trained model to those from the dropout deterministic sampler with varying prior covariance values ($P_{\text{prior}} = 0.01 \operatorname{I}, \operatorname{I},$ and $5 \operatorname{I}$) for RA test sentences (model analysis). The results show no significant difference in mean activations for $P_{\text{prior}} = 0.01 \operatorname{I}$ ($p > .001$), while significant differences are observed for larger prior covariances ($p < .001$).}
    \label{tab:paired_vary_prior}
    \end{center}
\end{table}
As reported in Table \ref{tab:paired_vary_prior}, the difference between output activations obtained with MLE-trained and Bayesian SG model were  statistically insignificant (\(p>0.001\)) only with a too small prior covariance (\(P_{\text{prior}} = 0.01 \operatorname{I}\)), while larger priors (\(P_{\text{prior}}=\operatorname{I}\) and \(5 \operatorname{I}\)) yielded significant differences (\(p<0.001\)). It was observed that with a very small prior covariance, the dropout deterministic sampler converged too quickly and was closer to the MLE estimates. In contrast, with a larger prior covariance, the sampler required more time steps to converge to the posterior but provided meaningful uncertainty estimates.

\bibliographystyle{elsarticle-harv} 
\bibliography{lib}
\end{document}